%% file: main.tex
\documentclass[journal]{IEEEtran}
\usepackage{amsmath,amsfonts}
\usepackage{algorithmic}
\usepackage{algorithm}
\usepackage{array}
\usepackage{textcomp}
\usepackage{stfloats}
\usepackage{url}
\usepackage{verbatim}
\usepackage{graphicx}
\usepackage{cite}
\usepackage{amssymb}
\usepackage{amsmath}
\usepackage{booktabs}
\usepackage{multirow}
\usepackage{caption}
\usepackage{subcaption}
\usepackage{xcolor}
\begin{document}

\title{Energy and Memory-Efficient Federated Learning with Ordered Layer Freezing}

\author{Ziru Niu, Hai Dong,~\IEEEmembership{Senior Member,~IEEE,} A. K. Qin,~\IEEEmembership{Fellow,~IEEE}, Tao Gu,~\IEEEmembership{Fellow,~IEEE} and Pengcheng Zhang,~\IEEEmembership{Member,~IEEE}
\thanks{Ziru Niu and Hai Dong (Corresponding Author) are with the School of Computing Technologies, RMIT University, Melbourne, VIC 3000, Australia. E-mail: ziru.niu@student.rmit.edu.au, hai.dong@rmit.edu.au.}
\thanks{A. K. Qin is with the Department of Computing Technologies, Swinburne University of Technology, Hawthorn, VIC 3122, Australia. E-mail: kqin@swin.edu.au.}
\thanks{Tao Gu is with the Department of Computing, Macquarie University,
Sydney, New South Wales, Australia. E-mail: tao.gu@mq.edu.au.}
\thanks{Pengcheng Zhang is with the Key Laboratory of Water Big Data Technology of Ministry of Water Resources \emph{and} the College of Computer and Software, Hohai University, Nanjing, China. E-mail: pchzhang@hhu.edu.cn.}
}

\markboth{IEEE TRANSACTIONS ON MOBILE COMPUTING}%
{Shell \MakeLowercase{\textit{et al.}}: A Sample Article Using IEEEtran.cls for IEEE Journals}


\maketitle

\begin{abstract}
Federated Learning (FL) has emerged as a privacy-preserving paradigm for training machine learning models across distributed edge devices in the Internet of Things (IoT). By keeping data local and coordinating model training through a central server, FL effectively addresses privacy concerns and reduces communication overhead. However, the limited computational power, memory, and bandwidth of IoT edge devices pose significant challenges to the efficiency and scalability of FL, especially when training deep neural networks. Various FL frameworks have been proposed to reduce computation and communication overheads through dropout or layer freezing. However, these approaches often sacrifice accuracy or neglect memory constraints. To this end, in this work, we introduce Federated Learning with Ordered Layer Freezing (FedOLF). FedOLF consistently freezes layers in a predefined order before training, significantly mitigating computation and memory requirements. To further reduce communication and energy costs, we incorporate Tensor Operation Approximation (TOA), a lightweight alternative to conventional quantization that better preserves model accuracy. Experimental results demonstrate that over non-iid data, FedOLF achieves at least 0.3\%, 6.4\%, 5.81\%, 4.4\%, 6.27\% and 1.29\% higher accuracy than existing works respectively on EMNIST (with CNN), CIFAR-10 (with AlexNet), CIFAR-100 (with ResNet20 and ResNet44), and CINIC-10 (with ResNet20 and ResNet44), along with higher energy efficiency and lower memory footprint.
\end{abstract}


\input{intro}
\input{literature}
\input{methodology}
\input{convergence}
\input{experiment}

\section{Conclusion}\label{sec:con}
This paper proposed Federated Learning with Ordered Layer Freezing (FedOLF), an efficient FL framework where edge devices only train the top-level layers of the model to accommodate resource constraints. The OLF strategy can minimize the backpropagation path length and the gradient error, which significantly reduces both the theoretical and practical memory requirements, while maintaining accuracy. We also enhance FedOLF with the Tensor Operation Approximation (TOA) technique, further alleviating energy consumption and memory footprint with less accuracy sacrifice. In the future, we aim to adapt FedOLF to dynamic networks where the system capacity of clients may vary, and enhance the engagement of FedOLF in IoT applications such as mobile edge networks and video surveillance. \par

\section*{Acknowledgments}
This work is funded by the Australian Research Council under Grant No. DP220101823, DP200102611, and LP180100114, National Natural Science Foundation of China under Grant No.62272145 and No. U21B2016, and Henan Provincial Key Research and Development Program
under Grant No. 251111210500.

 
%

\bibliography{reference.bib}{}
\bibliographystyle{IEEEtran}

 
\vspace{-30pt}

\begin{IEEEbiography}[{\includegraphics[width=1in,height=1.25in,clip,keepaspectratio]{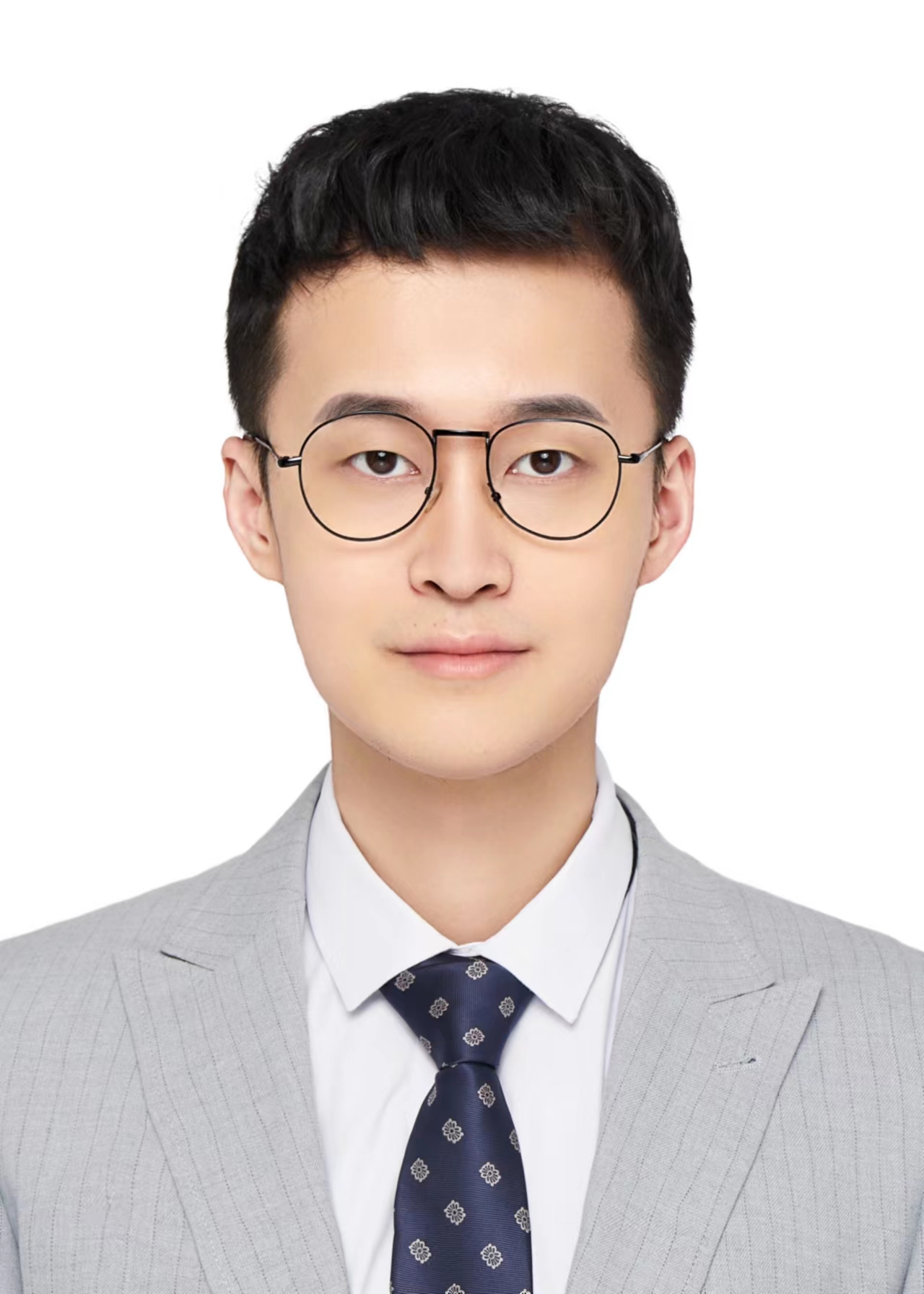}}]{Ziru Niu}
received the bachelor's degree and a Master degree in 2019 and 2021 respectively, both from the University of Melbourne, VIC, Australia. He is currently a PhD candidate from the School of Computing Technologies, RMIT University, Melbourne, VIC, Australia. His research interests include federated learning, edge computing, the Internet of Things, and trust management systems.
\end{IEEEbiography}
 \vspace{-30pt}
\begin{IEEEbiography}[{\includegraphics[width=1in,height=1.25in,clip,keepaspectratio]{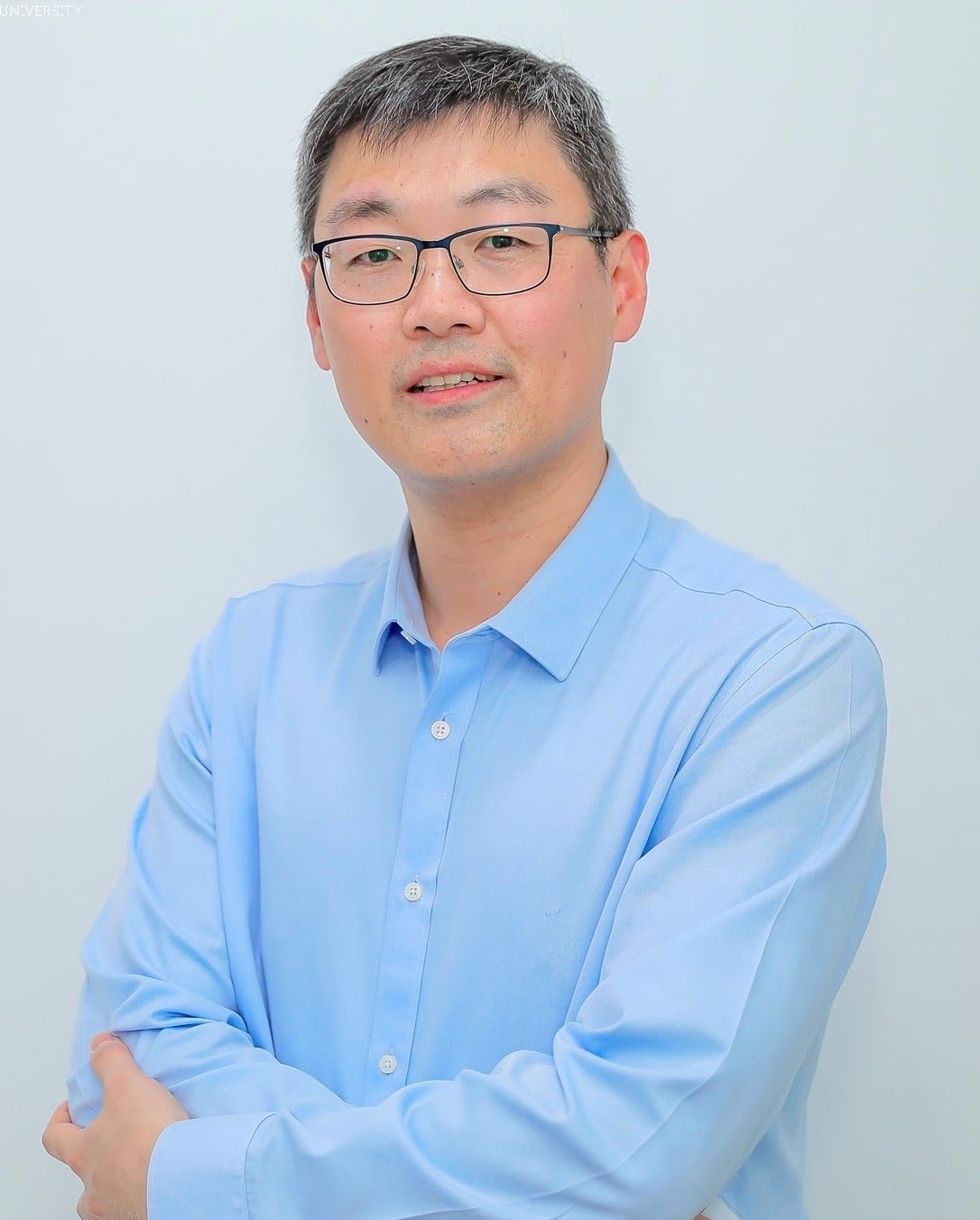}}]{Hai Dong} (Senior Member, IEEE) 
received a Ph.D. degree from Curtin University, Perth, Australia, and a Bachelor's degree from Northeastern University, Shenyang, China. He is currently an Associate Professor at the School of Computing Technologies, RMIT University, Melbourne, Australia. 
His primary research interests include Cloud/Edge Computing, Edge Intelligence, Blockchain, Cyber Security, and AI Security. 
\end{IEEEbiography}
\vspace{-30pt}
\begin{IEEEbiography}[{\includegraphics[width=1in,height=1.25in,clip,keepaspectratio]{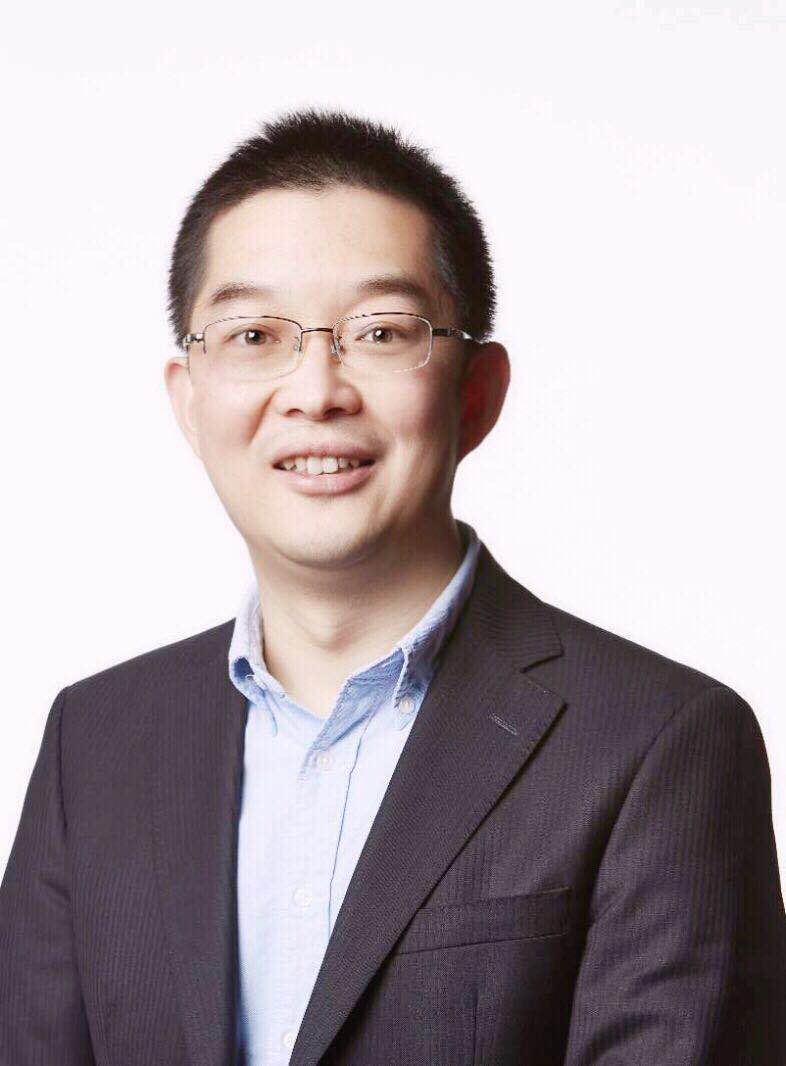}}]{A. K. Qin} (Fellow, IEEE) received the B.Eng. degree in Automatic Control from Southeast University, Nanjing, China, in 2001, and the Ph.D. degree in Computer Science and Engineering from Nanyang Technology University, Singapore, in 2007. 
He joined Swinburne University of Technology, Hawthorn, VIC, Australia, in 2017, where he is now a Professor.
His major research interests include machine learning, evolutionary computation, computer vision, remote sensing, services computing, and edge computing. 

\end{IEEEbiography}
\vspace{-30pt}
\begin{IEEEbiography}[{\includegraphics[width=1in,height=1.25in,clip,keepaspectratio]{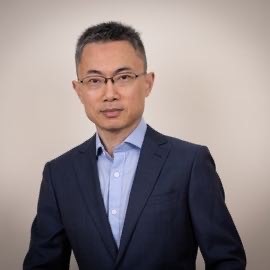}}]{Tao Gu} (Fellow, IEEE) is currently a Professor in Department of Computing at Macquarie University, Sydney. 
He obtained his Ph.D. in Computer Science from National University of Singapore, M.Sc. in Electrical and Electronic Engineering from Nanyang Technological University, and B.Eng. in Automatic Control from Huazhong University of Science and Technology. 
His current research interests include Internet of Things, Ubiquitous Computing, Mobile Computing, Embedded AI, Wireless Sensor Networks, and Big Data Analytics. He is a Distinguished Member of the ACM.
\end{IEEEbiography}
\vspace{-30pt}
\begin{IEEEbiography}[{\includegraphics[width=1in,height=1.25in,clip,keepaspectratio]{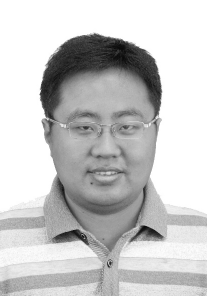}}]{Pengcheng Zhang} (Member, IEEE) received the Ph.D. degree in computer science from Southeast University in 2010. He is currently a full professor in the College of Computer and Software, Hohai University, Nanjing, China. His research interests include software engineering, service computing. He co-authored more than 70 peer-reviewed conference and journal papers, and has served as technical program committee member on various international conferences. He is a member of the IEEE.

\end{IEEEbiography}

\vfill

\end{document}

%% file: intro.tex
\section{Introduction}
\IEEEPARstart{E}{dge} devices have become an indispensable character in the Internet of Things (IoT) to offer intelligent services to end customers in various applications, like face recognition \cite{iotface} and audio analysis \cite{iotaudio}. These applications usually crave a deep learning model to process the complex data, such as image and voice \cite{goodfellow2016deep}. However, the decentralized topology of IoT networks brings significant challenges to the deployment of deep learning, as traditional machine learning approaches require edge devices to upload their data to a remote server for training \cite{flrce, hermes}. 
In the era of big data, massive data are generated by edge devices, which will introduce overwhelming network stress in transmission \cite{pa}. 
Additionally, transmitting highly sensitive data may cause severe privacy concerns due to the high possibility of data leakage in transmission \cite{hermes}.

As a solution, \textbf{Federated Learning (FL)} \cite{fedavg} has gained significant traction in IoT to process decentralized data and provide privacy-preserving services to clients
\cite{maqos, autofed, health_survey}. FL is a decentralized machine learning paradigm enabling edge devices to train a global model collaboratively while maintaining data privacy. Orchestrated by a central server, edge devices train deep learning models locally without sharing any personal data, and exchange the model parameters with the server. Then the server aggregates the received parameters to build a single global model, and broadcasts the model to all devices for their local tasks. Abstaining from data transmission, FL markedly alleviates the risk of data leakage and addresses privacy concerns. Moreover, FL significantly reduces the communication overhead, as the magnitudes of model parameters are usually lightweight compared with raw data \cite{fedavg}. \par

However, the heterogeneous nature of client devices poses a challenge due to varying system capacities. In real-world IoT environments, FL clients consist of diverse edge devices, and exhibit various configurations in terms of processor, battery, bandwidth, and memory. Resource-constrained devices with limited processor and bandwidth capabilities face difficulties in training and transmitting deep neural networks, leading to straggling, low quality-of-service, and excessive computation and communication costs. Moreover, devices with limited memory capacity may be unable to handle memory-intensive neural networks, thus being excluded from FL with severe information loss. Therefore, addressing the issue of resource constraints is crucial for the successful application of FL in IoT systems \cite{flRCsurvey, slt}. \par

Several studies have been proposed to address resource constraints through techniques such as \textbf{dropout} \cite{randdrop, fjord, heterofl, depthfl} and \textbf{layer freezing} \cite{slt, cocofl}. These methods involve training a subset of the global model with reduced requirements on computing resources, bandwidth, and memory on edge devices. Specifically, dropout involves pruning a fraction of the global model and sending the remaining sub-model to clients for training. However, it may significantly degrade accuracy on non-independently and identically distributed (non-iid) local data. In such settings, data importance among clients may vary, and training an underparameterized sub-model for an important client with data resembling the global distribution may not sufficiently capture knowledge from local data, leading to decreased accuracy of the global model \cite{cocofl, feddyn}. \par

Instead of sub-models, layer freezing involves sending the full global model to all devices and allowing resource-constrained devices to freeze some layers during training. For example, CoCoFL \cite{cocofl} allows clients to randomly train certain layers while freezing the remaining, while SLT \cite{slt} enables clients to sequentially train each layer in a bottom-up manner, with other layers partially frozen. Compared to dropout, layer freezing is more resilient to non-iid data by preserving the full model architecture on each client \cite{cocofl}. However, layer freezing introduces heavy communication overhead since the global model must be transmitted to clients. \par

\begin{figure}[ht]
    \centering
    \includegraphics[scale=0.6]{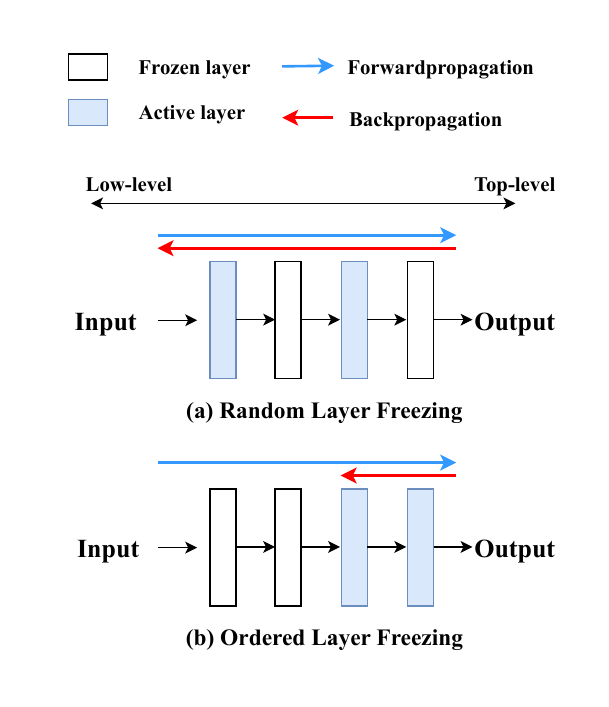}
 \caption{A comparison between \textbf{(a)} Random Layer Freezing and \textbf{(b)} Ordered Layer Freezing. The former requires more memory space to pass the gradient information back towards low-level active layers.}
 \label{lf_compare} 
\end{figure}

\textit{In addition, existing layer freezing techniques, such as \cite{cocofl} (random layer freezing) and \cite{slt} (top-first layer freezing), require an unexpectedly large memory usage for training, as they neglect the fact that the top frozen layers must store extra information in backpropagation, which might be a bottleneck for devices with limited memory capacity.} For example, Figure \ref{lf_compare} illustrates a comparison between two training modules: (a) random layer freezing and (b) ordered layer freezing. For random layer freezing (Figure  \ref{lf_compare}(a)), the top layers need to store the required information (i.e. activations and gradients) to compute the gradient of the bottom active layers, constituting a longer backpropagation path with higher memory usage. Comparatively, in ordered layer freezing (Figure \ref{lf_compare}(b)), low-level frozen layers will not store any gradient or activation as they do not participate in the gradient computation of the top active layers. Consequently, ordered layer freezing obtains a shorter backpropagation path with less memory consumption. \par

\begin{figure}[ht]
    \centering
    \includegraphics[scale=0.5]{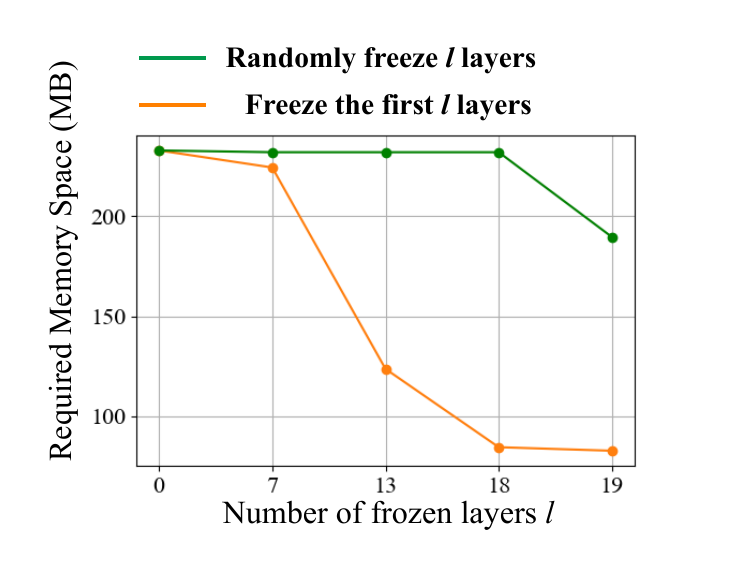}
 \caption{The required memory space of random and ordered layer freezing (Model: ResNet20, Dataset: CIFAR-100).}
 \label{lf_memory} 
\end{figure}

To validate this analysis, we implement these two layer-freezing strategies using ResNet20 \cite{resnet} with the CIFAR-100 dataset \cite{cifar}, and measure their \textit{required memory space} using the {\fontfamily{qcr}\selectfont TORCH.CUDA.MAX\underline{\hspace{1mm}}MEMORY\underline{\hspace{1mm}}ALLOCATED} function in PyTorch \cite{cudamemory}. As depicted in Figure \ref{lf_memory}, random layer freezing does not effectively alleviate the memory requirement in practice. The reason is that, for random layer freezing, the memory requirement is acknowledged as the maximum memory usage in the worst case, where the first bottom layers are active (as Figure \ref{lf_compare}(a) shows). Although the number of frozen layers increases, the oversized activation maps still have to be stored across all frozen layers for backpropagation, accounting for a dominant memory usage \cite{slt}. Contrarily, by shortening the path of backpropagation, ordered layer freezing notably mitigates the required memory space for training.
\par

Motivated by the limitations of existing methods, we introduce a practically efficient FL framework named \textbf{Federated Learning with Ordered Layer Freezing (FedOLF)}. In FedOLF, resource-constrained devices consistently freeze the low-level layers while training the remaining top-level layers. This approach substantially reduces the computation overhead and memory requirements of training, by shortening the gradient backpropagation path as illustrated in Figure \ref{lf_compare}(b). Additionally, we empirically observe that the gradient loss resulting from low-level frozen layers tends to diminish as training moves forward to top-level layers, which helps FedOLF maintain accuracy. Furthermore, we combine FedOLF with the \textbf{Tensor Operation Approximation (TOA)} scheme \cite{tensorapprox} to reduce the communication cost. Instead of the full global model, clients receive a sparsified approximation of the frozen layers with the full active layers from the server during communication. Unlike conventional quantization methods, TOA minimally impacts training and significantly preserves model accuracy. The contributions of this paper are summarized as follows:
\begin{itemize}
    \item We provide an insightful analysis discovering the limitations of existing layer freezing methods in addressing memory limitations in practice (see Figures \ref{lf_compare} and \ref{lf_memory}).  
    \item Based on the shortcomings of prior works, we introduce FedOLF, an efficient FL framework addressing memory limitation by allowing resource-constrained devices to train partial top-level layers of the global model (Section \ref{subsec:fedolf}). We also provide a convergence analysis of FedOLF in non-convex settings (Section \ref{sec:converge}). 
    \par
     \item We combine FedOLF with the Tensor Operation Approximation (TOA) framework to further reduce communication and memory costs (Section \ref{subsec:toa}). Compared with traditional quantization methods, TOA largely preserves the accuracy of FedOLF. To the best of our knowledge, this paper is among the first studies to apply TOA to reduce the memory and communication costs of FL.
    \par
     \item We evaluate FedOLF on EMNIST (with CNN), CIFAR-10 (with AlexNet), CIFAR-100 and CINIC-10 (with ResNet20 and ResNet44). Experimental results demonstrate that FedOLF outperforms the state-of-the-art by improving accuracy by at least 0.3\%, 6.4\%, 5.81\%, 4.4\%, 6.27\% and 1.29\% over non-iid client data, with higher energy efficiency and less memory consumption. 
\end{itemize}
The rest of the paper is organized as follows. Section \ref{lit_review} briefly reviews the related works. Section \ref{sec:methodology} describes the proposed FedOLF framework. Section \ref{sec:converge} provides a convergence analysis of FedOLF. Section \ref{sec:exp} evaluates the performance of FedOLF. Section \ref{sec:con} concludes this paper and lists future directions. 

%% file: literature.tex
\section{Related Work}\label{lit_review}

In this section, we briefly introduce the state-of-the-art related to efficient federated learning and federated learning on resource-constrained devices.

\subsection{Efficient Federated Learning} 
This stream of research aims to alleviate the computational and communication costs associated with FL. Various approaches have been proposed to enhance computation efficiency, such as FedProx \cite{fedprox}, FedParl \cite{fedparl}, and PyramidFL \cite{pyramid}, which reduce client training epochs to mitigate the overall computation costs. However, these works overlook communication efficiency and still require clients to transmit the entire model in FL. \par

To improve communication efficiency, FedCOM \cite{fedcom}, FetchSGD \cite{fetchsgd}, and STC \cite{flstc} reduce the size of transmitted parameters through message compression. FedSL \cite{fedsl}, FedOBD \cite{fedobd}, FedNew \cite{fednew}, Fedproto \cite{fedproto}, and DS-FL \cite{dsfl} advocate for transmitting lightweight replacement messages, such as logits and prototypes, instead of the full global model. These methods often focus on the singular aspect of communication efficiency and fail to address the computation bottlenecks of edge devices.  \par

To simultaneously achieve computation and communication efficiency, adaptive dropout \cite{hermes, prunefl, adadrop, fedmask} and progressive parameter freezing \cite{alf, apf, yibowu} offer a more comprehensive approach by enabling clients to train and transmit a subset of the entire model (i.e., a \textit{sub-model}), thereby achieving both computation and communication efficiency. In these works, clients first train the entire model for a few rounds, and assess the importance of each neuron/layer based on some heuristics (e.g. \(l-\)1 and \(l-\)2 norms). Accordingly, clients prune/freeze the unimportant parameters with the least heuristics, and train/transmit the remaining \textit{sub-model} with fewer computation and communication costs. Nevertheless, the strategy of separately executing sub-model extraction and training can be time-consuming. Besides, these methods overlook memory constraints, as clients must prune or freeze the unimportant parameters to generate sub-models, a process that requires pre-training the full model locally. Although FIARSE \cite{fiarse} effectively accelerates training by jointly optimizing masks and parameters, it does not overcome memory constraints, because it employs a strategy of masking parameters based on their magnitudes, which requires full-model training. Lastly, FLrce \cite{flrce} mitigates overall computation and communication costs by reducing FL iterations with an early-stopping mechanism, but it still overlooks memory constraints by performing full-model training over clients. \par

Furthermore, none of the aforementioned works adequately account for memory constraints on devices, as they typically involve full-model training on all clients regardless of their memory capacity. 

\subsection{Federated Learning on Resource-Constrained Devices} 
The primary distinction between efficient FL and resource-constrained FL lies in the latter's consideration of devices with limited resources, such as memory space or bandwidth support, which are unable to train or transmit the entire model. To tackle this challenge, \cite{randdrop, fjord, depthfl, heterofl, adaptivefl, scalefl, nefl} introduce the concept of \textit{sub-models}, which contain fewer parameters and can be trained and transmitted by resource-constrained clients. Specifically, Feddrop \cite{randdrop} employs random neuron pruning, FjORD \cite{fjord}, HeteroFL \cite{heterofl} and AdaptiveFL \cite{adaptivefl} adopt a right-to-left approach for neuron pruning. DepthFL \cite{depthfl} and NeFL \cite{nefl} respectively employ top-first and medium-first layer pruning. Lastly, the hybrid ScaleFL method \cite{scalefl} combines bottom-first layer pruning and left-most neuron pruning. 
\par
Unlike adaptive dropout, these works execute dropout at the server side, eliminating the need for clients to pre-train a full model. However, these methods are susceptible to non-iid data among clients, as training small sub-models on crucial clients may not capture sufficient knowledge to construct an accurate global model. \par

In contrast, CoCoFL \cite{cocofl} and SLT \cite{slt} advocate for maintaining the full model architecture on all clients while freezing certain layers on resource-constrained devices. CoCoFL randomly freezes layers within the local model, whereas SLT partially freezes top-level layers and sequentially trains all layers from the bottom. The frozen layers remain untrained and untransmitted to enhance computation and communication efficiency. However, these approaches lead to increased memory usage, particularly in the case of frozen top-level layers, which consume significant memory space to transmit gradient information backward, as illustrated in Figure \ref{lf_compare}. Similarly, FedPT \cite{fedpt} proposes to freeze the blocks with the most parameters in training, which may not be memory-efficient as the largest blocks do not necessarily contain the bottom layers. Moreover, TinyFEL \cite{tinyfel} proposes to block the backpropagation path towards the bottom layers in training for computation and memory efficiency. However, this approach neglects the fact that in practical implementations (e.g. Pytorch), while remaining active in forward propagation, the bottom layers may still generate massive activations in training, resulting in high memory consumption (see Section \ref{sec:tinyfel} for details).

In summary, none of the aforementioned works can simultaneously achieve computation, communication, and memory efficiency without sacrificing accuracy. To this end, we expect a comprehensive FL framework that overcomes these challenges.

%% file: methodology.tex
\begin{figure}[ht]
    \centering
    \includegraphics[scale=0.6]{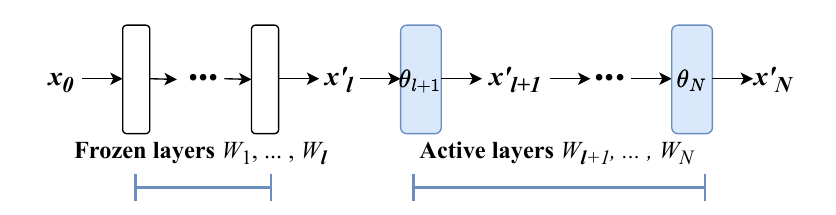} 
 \caption{During training, the \(l\) frozen layers will generate a feature representation \(\boldsymbol{x'}_{l}\) that diverges from the true \(\boldsymbol{x}_{l}\). Affected by \(\boldsymbol{x'}_{l}\), the following active layers also generate inaccurate representations.}
 \label{img:vanish} 
 \end{figure}

\begin{figure*}[htb]
    \centering
    \includegraphics[scale=0.4]{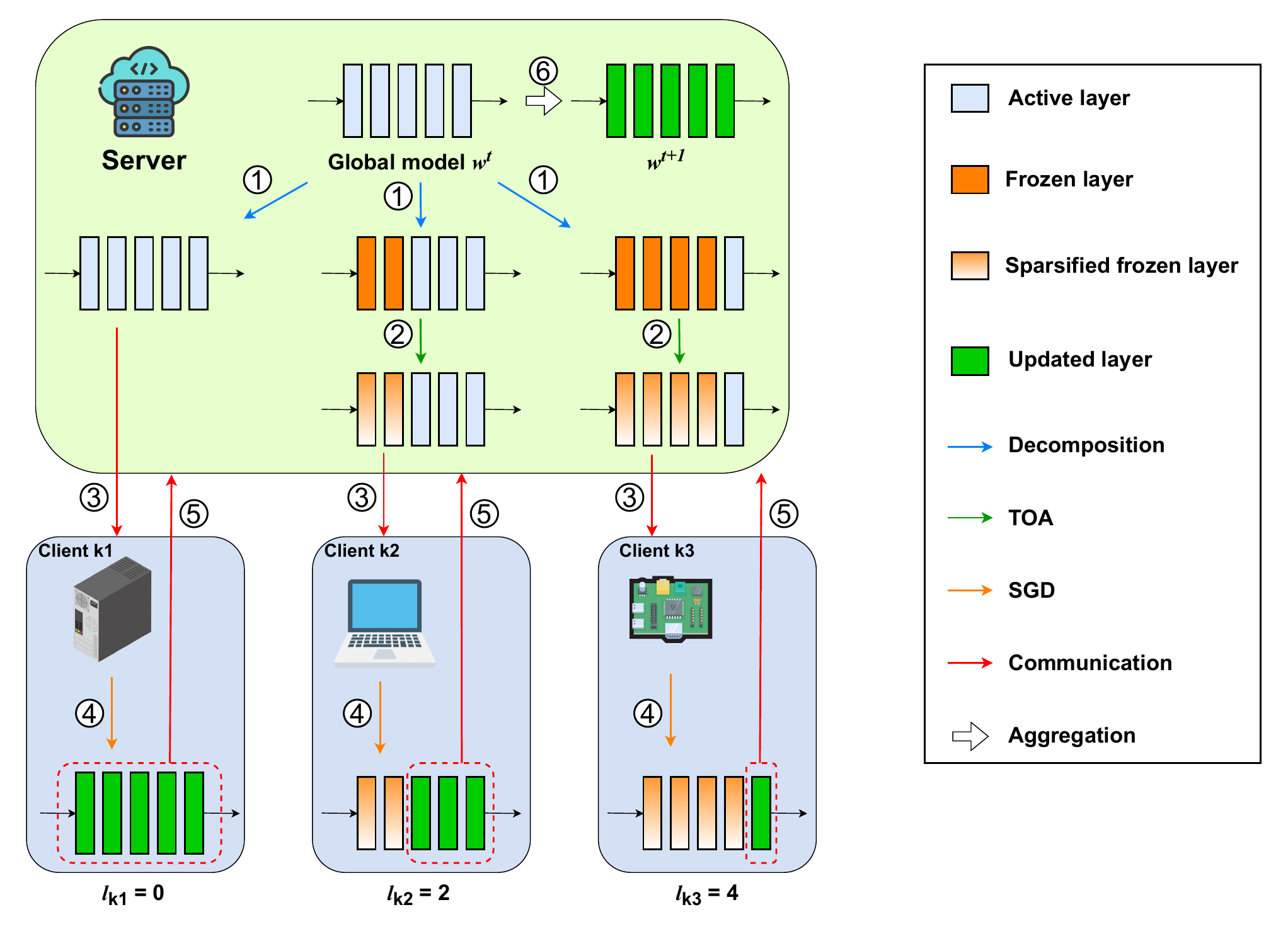}
 \caption{A high-level flow chart demonstrating how FedOLF works. For simplicity, we use an exemplary network consisting of a global model with five layers, and three clients \(k_{1}, k_{2}\) and \(k_{3}\) with three degrees of system capacity.}
 \label{img:olfdemo} 
\end{figure*}

\section{Methodology}\label{sec:methodology}
Motivated by the limitations of existing works (as shown in Figures \ref{lf_compare} and \ref{lf_memory}), we propose \textbf{FedOLF} (Federated Learning with Ordered Layer Freezing). The key idea of FedOLF is to let resource-constrained devices consistently freeze low-level layers in training, thereby freeing frozen layers from storing additional information in backpropagation, such as activations and gradients \cite{slt}. This module enables FedOLF to minimize the length of the backpropagation path and effectively reduce memory consumption, as shown in Figure \ref{lf_compare}(b) and Figure \ref{lf_memory}. Additionally, we combine FedOLF with the \textbf{Tensor Operation Approximation (TOA)} technique to further reduce the communication cost. 

\subsection{Preliminaries}
Given a network with one server and \(K\) devices (clients), and a global model \(w\) stored on the server side, the goal of FL is to optimize the following problem:
\begin{equation}
\begin{split}
    & \mathop{\min}_{w} f(w) := \mathbb{E}[f_{k}(w)] := \sum_{k=1}^{K}\frac{n_{k}}{n}(f_{k}(w)) , \\
    &  f_{k}(w) := \frac{1}{n_{k}} \sum^{i=1}_{n_{k}} \mathcal{L}(w, (X_{i}, y_{i})) . \\
\end{split}
\end{equation}
\(f\), the global objective function, is a weighted average of all local objective functions \(f_{k}\) (\(1\leq k \leq K\)). For a client \(k\), the local objective function \(f_{k}\) is equivalent to the empirical risk over its personal dataset \(D_{k}\), \(n_{k} = |D_{k}|\) is the size of the local dataset and \(\mathcal{L}(w, (X_{i}, y_{i}))\) is the prediction loss of \(w\) over the \(i-\)th sample \((X_{i}, y_{i})\) in \(D_{k}\). \(n=\sum_{k=1}^{K}n_{k}\) is the total number of samples across all local datasets. Moreover, let \(N\) denote the total number of layers in the global model \(w\), and \(W_{l}\) represent the \(l-\)th layer with parameter \(\boldsymbol{\theta}_{l}\) (\(1\leq l \leq N\)). The layer \(W_{l}\) can be viewed as a function that takes the input feature representation \(\boldsymbol{x}_{l-1}\) from the previous layer, and outputs a new feature representation \(\boldsymbol{x}_{l}\), i.e. \(\boldsymbol{x}_{l} = W_{l}(\boldsymbol{x}_{l-1}, \boldsymbol{\theta}_{l})\). Specially, \(\boldsymbol{x}_{0} = X\) is the initial data sample, and \(\boldsymbol{x}_{N}=\boldsymbol{\hat{y}}\) is the model's final prediction. \par

As Figure \ref{img:vanish} shows, for a client \(k\), the architecture of model \(w\) can be decomposed into two components \(w_{F,k}\) and \(w_{A,k}\) such that \(w= w_{F,k} \cup w_{A,k}\). \(w_{F,k} = \{W_{1},...,W_{l_{k}}\}\) and \(w_{A,k}=\{W_{l_{k}+1},...,W_{N}\}\) are respectively the set of frozen and active layers. \(l_{k} \in \{0,1,...,N-1\}\) is the number of frozen layers in training whose value depends on \(k\)'s device capacity. For a powerful device that can train the entire model, we have \(l_{k} = 0\) and \(w_{F,k} = \varnothing \). \par

\begin{figure}[ht]
    \includegraphics[width=0.52\textwidth]{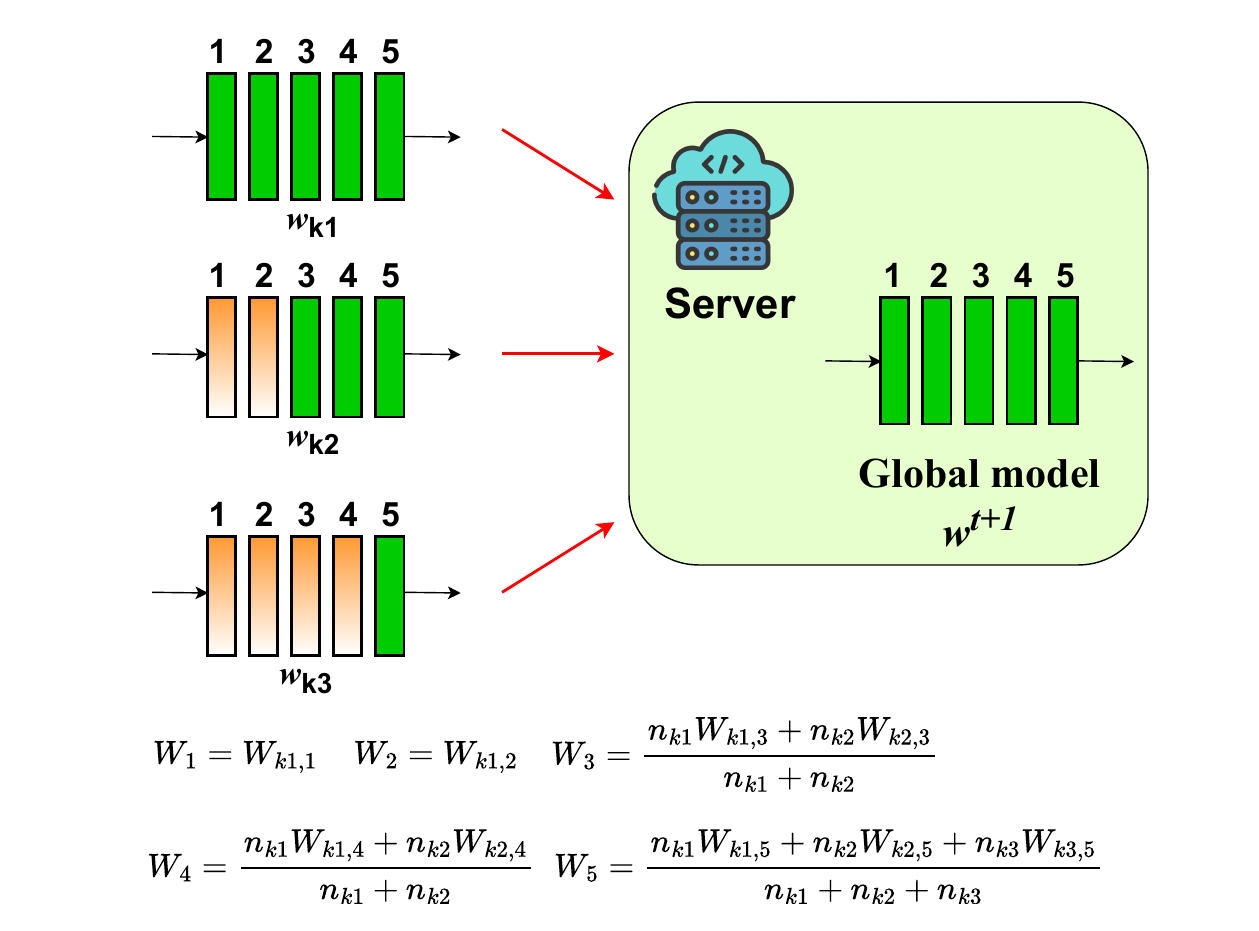} 
 \caption{An illustration of the layer-wise aggregation scheme in FedOLF (same as in \cite{cocofl, fedsl}). \(W_{l}\) and \(W_{k,l}\) respectively stand for the \(l\)-th layer in the global model and the local model of client \(k\). Here we have \(1\leq l \leq 5\) and \(k \in \{k_{1}, k_{2}, k_{3}\}\).}
 \label{img:lfagg} 
 \end{figure}

\subsection{FedOLF Overview}\label{subsec:fedolf}
The FedOLF framework is summarized as Figure \ref{img:olfdemo} shows. Specifically: \par
\textcircled{1}. At global iteration \(t\), for each selected client \(k\), the server decomes the global model \(w^{t}\) into two fractions based on \(l_{k}\), which are frozen layers \(w_{F,k} = \{W_{1},...,W_{l_{k}}\}\) and active layers \(w_{A,k}=\{W_{l_{k}+1},...,W_{N}\}\). \par

\textcircled{2}. For each selected \(k\), the server sparsifies the frozen layers using the TOA algorithm (see Section \ref{subsec:toa}) for communication efficiency. For powerful clients with the ability to perform full-model training, such as \(k_{1}\) in Figure \ref{img:olfdemo}, the set of frozen layers is empty and TOA is not applied. \par

\textcircled{3}. Each client downloads the full active layers and the sparsified frozen layers from the server. For clients with full-model training ability (e.g. \(k_{1}\) in Figure \ref{img:olfdemo}), the entire model will be downloaded.\par

\textcircled{4}. Client \(k\) locally trains all parameters in \(w_{A,k}^{t}\) by applying stochastic gradient descent (SGD) on dataset \(D_{k}\) according to Equation (\ref{eq:lf_sgd}):
\begin{equation}\label{eq:lf_sgd}
    w_{A,k}^{t+1} \gets w_{A,k}^{t} - \eta \nabla f'_{k} (w_{A,k}^{t})
\end{equation}
\(\eta\) is the learning rate and \(\nabla f'_{k}\) is a low-error-rate approximation of the gradient \(\nabla f_{k}\) in the case of layer freezing. 

\textcircled{5}. Once local training is completed, client \(k\) only sends the updated layers \(w_{A,k}^{t+1}\) to the server for communication efficiency. \par 

\textcircled{6}. After receiving the results from all participating clients, the server updates the global model using the same layer-wise aggregation strategy as in \cite{cocofl, fedsl} (see Figure \ref{img:lfagg}), then moves forward to round \(t+1\) and restarts from step \textcircled{1}. The complete procedure of FedOLF is presented in Algorithm \ref{alg:fedolf}. \par

\begin{algorithm}[tb]
   \caption{The procedure of FedOLF}
   \label{alg:fedolf}
\begin{algorithmic}[1]
   \STATE {\bfseries Input:} maximum global iteration \(T\), clients \(C=\{1,...,K\}\) with numbers of frozen layers \(\{l_{1},...,l_{K}\}\), and initial global model \(w^{0}\), scale factor \(s\).
   \FOR {\(t = 1,2,...,T\)}
   \STATE Sample a set of participating clients \(C_{t} \subset C\).
   \STATE \textbf{for} every client \(k \in C_{t}\) \textbf{server does:}
   \STATE \hspace{5mm} Decompose \(w^{t}\) into \(w_{F,k}^{t}\) and \(w_{A,k}^{t}\) based on \(l_{k}\).
   \STATE  \hspace{5mm} \(\hat{w}_{F,k}^{t} \gets TOA(w_{F,k}^{t}, s, l_{k})\).  \hspace{10mm}   (see Alg. \ref{alg:toa})
   \STATE \hspace{5mm} Send \(\hat{w}_{F,k}^{t}\) and \(w_{A,k}^{t}\) to \(k\).
   \STATE \textbf{each} \(k \in C_{t}\) \textbf{in parallel does}:
   \STATE \hspace{5mm} \(w_{k}^{t} \gets \hat{w}_{F,k}^{t} \cup w_{A,k}^{t}\).
   \STATE \hspace{5mm} \textbf{for} local epochs \(1,...,E\):
   \STATE \hspace{8mm} \(w_{A,k}^{t+1} \gets w_{A,k}^{t} - \eta \nabla f'_{k} (w_{A,k}^{t})\).
   \STATE \hspace{5mm} Upload \(w_{A,k}^{t+1}\) to the server. 
   \STATE \textbf{for each layer \(W_{l} \in w^{t}\), server does: }
   \STATE \hspace{5mm} Update \(W_{l}\) using weighted average (see Fig. \ref{img:lfagg}).
   
   \ENDFOR
   \STATE \textbf{Return} \(w^{t}\)

\end{algorithmic}
\end{algorithm}

In step \textcircled{4}, with layer freezing, the layers in \(w_{F,k}^{t}\) will remain constant as training goes on, and will subsequently generate a straggling feature representation \(\boldsymbol{x}'_{l_{k}} = \boldsymbol{x}_{l_{k}} + \boldsymbol{\sigma}_{l_{k}}\) as Figure \ref{img:vanish} shows. \(\boldsymbol{x}_{l_{k}}\) is the true representation generated by \(w_{F,k}^{t}\) if it is non-freezing, and \(\boldsymbol{\sigma}_{l_{k}}\) is an error term representing the divergence between \(\boldsymbol{x}_{l_{k}}\) and \(\boldsymbol{x}'_{l_{k}}\). Feeding \(\boldsymbol{x}_{l_{k}}\) and \(\boldsymbol{x}'_{l_{k}}\) forward will respectively result in \(\nabla f_{k}\) and \(\nabla f'_{k}\). However, based on the study of \cite{coreset}, although the feature representations after layer \(W_{l}\) diverge from the true representations owing to the staleness of frozen layers, the representation error \(\boldsymbol{\sigma}_{l_{k}}\) and the gradient error \(\|\nabla f'_{k}(w) - \nabla f_{k}(w) \|\) are usually upper bounded. This important property ensures that the negative impact of layer freezing on FedOLF's performance is limited, enabling FedOLF to train an accurate global model among resource-constrained devices. 

Furthermore, according to \cite{fedsl, nofear}, low-level layers across various local models usually have higher degrees of Centered Kernal Alignment (CKA) similarity across different datasets \cite{cka}, which means that these layers contain substantial redundant information and may generate similar feature representations. Motivated by this insight, in FedOLF, a resource-constrained device \(k\) can "borrow" the highly-generalized low-level layers from other clients by downloading \(w_{F,k}^{t}\) from the server. Layers in \(w_{F,k}^{t}\) have been trained by more powerful clients in previous rounds, and can be directly employed by \(k\) during the forward propagation phase of training without incurring significant errors. \par

\subsection{FedOLF with Tensor Operation Approximation}\label{subsec:toa}
Furthermore, we propose an adapted Tensor Operation Approximation (TOA) framework \cite{tensorapprox} dedicated to reducing the communication cost in FedOLF. Instead of the entire global model \(w\), a client \(k\) downloads \(\hat{w}_{F,k}^{t}\) and \(w_{A,k}^{t}\) from the server, where \(\hat{w}_{F,k}^{t}\) is a sparsified approximation of the frozen layers \(w_{F,k}^{t}\) with fewer parameters. Unlike the initial TOA method that works on all layers, in this paper, the modified TOA works only on the frozen layers to ensure all active layers get fully trained. For illustration, let \(H_{q}\) denote the number of tensors in a frozen layer \(W_{q}\), where a tensor is a filter or neuron if \(W_{q}\) is a convolution or fully connected layer, respectively. \par
\begin{figure}[ht]
    \centering
    \includegraphics[scale=0.4]{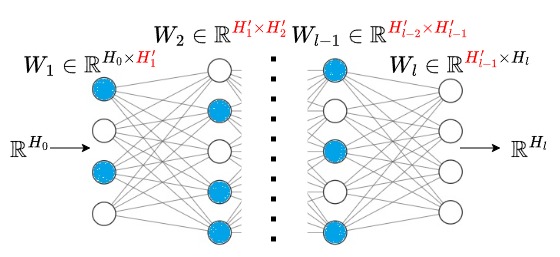}
    \caption{Within each frozen fully-connected layer \(W_{q}\) (\(1\leq q < l\)) containing \(H_{q}\) neurons, a subset \(W'_{q}\) (blue neurons) is derived by sampling \(H'_{q} = \lfloor sH_{q} \rfloor\) neurons of the layer. Consequently, the approximation of \(w_{F,k}^{t}\), represented as \(\hat{w}_{F,k}^{t}\) is \(\hat{w}_{F,k}^{t} = W'_{1}\cup......\cup W'_{l-1} \cup W_{l} \).}
    \label{fig:ta_demo}
\end{figure}
For example, Figure \ref{fig:ta_demo} shows how TOA is applied on a fully connected neural network with \(l\) frozen layers. For every layer \(W_{q}\) (\(1\leq q < l\)), except for the last frozen layer, the server samples \(\lfloor sH_{q} \rfloor\) tensors from the layer and sends this subset of tensors to client \(k\). \(s\) (\(0 < s \leq 1\)) is a scaling factor that determines the trade-off degree between accuracy and communication efficiency, with \(s=1\) representing that no TOA is applied. Moreover, TOA is not performed on the last frozen layer, as shown in Figure \ref{fig:ta_demo}, so that the dimensions of the representation output \(\boldsymbol{x'}_{l}\) and the following active layers remain unchanged. Based on the study of \cite{tensorapprox}, we apply a weighted sampling strategy on TOA.
With this strategy, TOA selects a tensor  \(\boldsymbol{Z}_{j,q}\) (\(1\leq j \leq H_{q}\)) within a frozen layer \(W_{q}\) 
  with probabilities proportional to their \textit{Frobenius} norms:
\begin{equation}\label{eq:taws}
    \mathbb{P}(\boldsymbol{Z}_{j,q} \in W'_{q}) = \frac{\|\boldsymbol{Z}_{j,q}\|_{F}}{\sum_{j=1}^{H_{q}} \|\boldsymbol{Z}_{j,q}\|_{F}}.
\end{equation}

In this case, the approximation error \(\mathbb{E}[\|\boldsymbol{x'}_{l} - \boldsymbol{x'}_{l,TOA}\|^{2}]\) will be minimized, where \(\boldsymbol{x'}_{l,TOA}\) and \(\boldsymbol{x'}_{l}\) are respectively the representation outputs by the TOA-sparfisied layer \(W'_{l}\) and the original frozen layer \(W_{l}\). The TOA technique significantly reduces the downstream communication cost in FedOLF by approximately \(O(s^{2})\). The procedure of TOA is shown in Algorithm \ref{alg:toa}. \par

\begin{algorithm}
\caption{The procedure of Tensor Operation Approximation}
\label{alg:toa}
\begin{algorithmic}[1]
\STATE {\bfseries Input:} set of frozen layers \(w_{F}\), scaling factor \(s\), number of frozen layers \(l_{k}\) (\(l_{k}\geq 2\)):
\FOR{every layer \(W_{q} \in w_{F}, 1\leq q \leq l_{k}-1\)}
\STATE \(H_{q} \gets len(W_{q})\).
\STATE \(W^{'}_{q} \gets\) $WeightedSample$(candidates=\(\{Z_{j,q}\}_{j=1}^{H_{q}}\),
\STATE \hspace{30mm} weights=\(\{\mathbb{P} (Z_{j,q} \in W^{'}_{q})\}_{j=1}^{H_{q}}\),
\STATE  \hspace{30mm} number=\(\lfloor sH_{q}\rfloor\)).
\ENDFOR
\STATE \textbf{Return} \(\hat{w}_{F} := W'_{1}\cup ...... \cup W'_{l_{k}-1}\cup W_{l_{k}}\)  
\end{algorithmic}
\end{algorithm}

An overview of the mathematical notations is presented by Table \ref{tab:notations}.
    
\begin{table}[ht]
    \centering
    \begin{tabular}{c|c}
         \hline
         \textbf{Notations} & \textbf{Semantics} \\
         \hline
         \(K\) & Total number of clients. \\
         \(k\) & The \(k\)-th client. \\
         \(T\) & Total number of global iterations. \\
         \(t\) & The \(t\)-th round of global communication. \\
         \(C\) & Universal set of all clients. \\
         \(C_{t}\) & The set of selected clients at round \(t\). \\
         \(D_{k}\) & Local dataset of client \(k\). \\
         \(n_{k}\) & Number of samples in the local dataset of client \(k\). \\
         \(n\) & Total number of samples across all clients. \\
         \(f\) & Global objective function. \\
         \(f_{k}\) & Local objective function of client \(k\). \\
         \(\nabla f_{k}\) & Real gradient of client \(k\) with no OLF applied. \\
         \(\nabla f'_{k}\) & Approximation of \(\nabla f_{k}\) under OLF. \\
         \(w^{t}\) & Global model at round \(t\). \\
         \(w_{k}^{t}\) & Local model of client \(k\) at round \(t\).\\
         \(N\) & Total number of layers in the global model. \\
         \(W_{l}\) & The \(l\)-th layer in the global model. \\
         \(W'_{l}\) & The \(l\)-th layer after the sparsification of TOA. \\
         \(\boldsymbol{\theta}_{l}\) & The parameters of layer \(W_{l}\). \\
         \(W_{k,l}\) & The \(l\)-th layer in the local model of client \(k\). \\
         \(l_{k}\) & Number of frozen layers in client \(k\)'s local model. \\
         \(\boldsymbol{x}_{l_{k}}\) & Original representation of the \(l_{k}\)-th layer of client \(k\). \\
         \(\boldsymbol{x'}_{l_{k}}\) & Representation of the last frozen layer of client \(k\). \\
         \(\boldsymbol{x'}_{l, TOA}\) & Representation output of a TOA-sparsified layer \(W'_{l}\). \\
         \(w_{A,k}^{t}\) & Set of active layers in client \(k\)'s local model at round \(t\). \\
         \(w_{F,k}^{t}\) & Set of frozen layers in client \(k\)'s local model at round \(t\). \\
         \(\hat{w}_{F,k}^{t}\) & Sparsified approximation of \(w_{F,k}^{t}\) after applying TOA. \\
         \(s\) & Scaling factor of TOA. \\
         \(H_{q}\) & Number of tensors in the \(q\)-th layer. \\
         \(H'_{q}\) & Number of tensors in the TOA-sparsified \(q\)-th layer. \\
         \(\boldsymbol{Z}_{j,q}\) & The \(j\)-th tensor in the \(q\)-th layer. \\
         \(\eta\) & Learning rate. \\
         \(E\) & Local epoch. \\
         \hline
    \end{tabular}
    \caption{Summary of mathematical notations.}
    \label{tab:notations}
\end{table}

%% file: convergence.tex
\section{Convergence Analysis}\label{sec:converge}
In this section, we analyze the convergence results for FedOLF on non-convex smooth objective functions. We do not require the objective function to be convex in the case of deep-learning neural networks \cite{scaffold}. 
\subsection{Assumptions and Theorems}
We first make the following assumptions: \par
\begin{itemize}
    \item \textbf{Assumption 1} (smoothness). \textit{The objective function \(f_{k}\) is \(L\)-smooth (\(L>0\)):}
    \begin{equation}\label{eq:smth}
    \forall w_{1}, w_{2}, \hspace{1mm} \|\nabla f_{k}(w_{1}) - \nabla f_{k}(w_{2}) \| \leq L \|w_{1} - w_{2}\|.
    \end{equation}
    \item \textbf{Assumption 2} (Bounded variance). \textit{The variance of local gradients to the global gradient is bounded with parameter \(\gamma\) (\(\gamma>0\)):}
    \begin{equation}
    \forall k,w, \hspace{1mm} \mathbb{E}(\|\nabla f_{k}(w)-\nabla f(w)\|^{2}) \leq \gamma^{2}.
    \end{equation}
\end{itemize}

Assumptions 1 and 2 are common assumptions that are made by extensive prior works, such as \cite{hermes, fedproto, fedprox, adadrop, scaffold, feddyn, longtail}. Furthermore, following \cite{coreset}, we can assume that the divergence of local gradient \(\|\nabla f_{k}' - \nabla f_{k} \|\) resulting from layer freezing is bounded: \par
\begin{itemize}
    \item \textbf{Assumption 3}. \textit{For any client \(k\), the divergence between the local gradient with and without layer freezing is upper bounded by \(D\) (\(D>0\)):} 
    \begin{equation}
    \forall k,w, \hspace{1mm} \|\nabla f'_{k}(w) - \nabla f_{k}(w) \|^{2} \leq D^{2}. 
    \end{equation}
\end{itemize}

Based on Assumptions 1-3, we derive the following theorems:
\begin{itemize}
    \item \textbf{Theorem 1}. \textit{When the learning rate \(\eta\) satisfies \(\frac{1}{L} < \eta < \frac{3}{2L}\), we have:}
    \begin{equation}
    \begin{split}
    f(w^{t+1}) - f(w^{t}) & \leq \frac{\eta}{2} (2\eta L -3) (\mathbb{E}[\|\nabla f(w^{t})\|])^{2} \\
    & +\eta D (\eta L - 1) \mathbb{E}[\|\nabla f (w^{t})\|] \\
    & + \frac{\eta}{2} (2\eta L\gamma^{2} - \gamma^{2} + \eta L D^{2} + 2\eta L D \gamma).  \\
    \end{split}
    \end{equation}
    \item \textbf{Theorem 2}. \textit{When the learning rate \(\eta\) satisfies \(\eta \leq \frac{1}{L}\), we have:}
\begin{equation}
    f(w^{t+1}) - f(w^{t}) \leq \frac{\eta}{2} \times (-\mathbb{E}[\|\nabla f(w^{t})\|^{2}] + D^{2} + \gamma^{2} + 2D\gamma).
\end{equation}
\end{itemize}

According to Theorem 1 and Theorem 2, when the learning rate is less than \(\frac{3}{2L}\), the objective function \(f\) continues to decrease until \(w^{t}\) converges to a \textit{\(\epsilon\)-critical point} where \(\|\nabla f(w^{t})\| \leq \epsilon\). Let \(w^{\epsilon}\) indicate the point where \(\|f(w^{\epsilon})\|) = \epsilon\), before reaching \(w^{\epsilon}\), for the global model \(w^{t}\), we continuously have \(\|\nabla f(w^{t})\| > \epsilon\). Subsequently, given an arbitrary initial model \(w^{0}\), the FedOLF algorithm is expected to converge to \(w^{\epsilon}\) in \(T = O(\frac{\|w^{\epsilon}-w^{0}\|}{\epsilon\eta})\) iterations. \par

In particular, when \(\frac{1}{L} < \eta < \frac{3}{2L}\), we have \(\epsilon = \epsilon_{1} = \frac{D(\eta L-1) + \sqrt{\eta D^{2}L+8\eta L \gamma^{2} + 6\eta DL\gamma+D^{2}-3\gamma^{2}}}{3-2\eta L}\). When \(\eta \leq \frac{1}{L}\), we have \(\epsilon = \epsilon_{2} = D + \gamma\). 

\subsection{Proof}
Since every \(f_{k}\) is \(L\)-smooth based on Assumption 1, \(f\) is also \(L\)-smooth, so that we have:
\begin{equation}
  f(w^{t+1}) - f(w^{t}) \leq \langle w^{t+1}-w^{t}, \nabla f(w^{t}) \rangle + \frac{L}{2} \|w^{t+1}-w^{t}\|^{2}. 
\end{equation}
In the setting of layer freezing, we have \(w^{t+1} = w^{t} - \eta \nabla f'(w^{t})\) and \(\nabla f'(w^{t}) = \mathbb{E}[\nabla f_{k}'(w^{t})]\). Therefore:
\begin{equation}\label{eq:pt1}
\begin{split}
    & f(w^{t+1}) - f(w^{t}) \\
    &  \leq -\eta \hspace{1mm} \langle  \hspace{1mm} \mathbb{E}[\nabla f'_{k}(w^{t})], \nabla f(w^{t})  \hspace{1mm} \rangle + \frac{L}{2}\|-\eta \mathbb{E}[\nabla f'_{k}(w^{t})] \|^{2} \\
    & = -\eta \mathbb{E} \hspace{1mm} [\langle  \hspace{1mm} \nabla f'_{k}(w^{t}), \nabla f(w^{t})  \hspace{1mm} \rangle] + \frac{L\eta^{2}}{2} \|\mathbb{E}[\nabla f'_{k}(w^{t})] \|^{2} \\
    & \leq -\eta \mathbb{E} \hspace{1mm} [\langle  \hspace{1mm} \nabla f'_{k}(w^{t}), \nabla f(w^{t})  \hspace{1mm} \rangle] + \frac{L\eta^{2}}{2}\mathbb{E}(\|\nabla f'_{k}(w^{t}) \|^{2}). \\
\end{split}
\end{equation}
Since \(\|\nabla f'_{k}(w^{t}) - \nabla f(w^{t})\|^{2} = \|\nabla f'_{k}(w^{t})\|^{2} -2 \hspace{1mm} \langle   \nabla f'_{k}(w^{t}), \nabla f(w^{t})  \hspace{1mm} \rangle + \|\nabla f(w^{t})\|^{2} \), Equation (\ref{eq:pt1}) can be written as:
\begin{equation}\label{eq:pt2}
\begin{split}
    & f(w^{t+1}) - f(w^{t}) \\
    & \leq \frac{\eta}{2}\mathbb{E}(\|\nabla f'_{k}(w^{t}) - \nabla f(w^{t})\|^{2} - \|\nabla f'_{k}(w^{t})\|^{2} - \|\nabla f(w^{t})\|^{2}) \\
    & + \frac{L\eta^{2}}{2}\mathbb{E}(\|\nabla f'_{k}(w^{t}) \|^{2}) \\
    & = \frac{\eta}{2} \mathbb{E}(\|\nabla f'_{k}(w^{t}) - \nabla f(w^{t})\|^{2}) \\
    & + \frac{\eta}{2}(\eta L - 1) \mathbb{E}[\|\nabla f'_{k}(w^{t}) \|^{2}] - \frac{\eta}{2} \mathbb{E}[\|\nabla f(w^{t}) \|^{2}] \\
    & =  \frac{\eta}{2} \mathbb{E}\|(\nabla f'_{k}(w^{t}) - \nabla f_{k}(w^{t}) \\
    & + \nabla f_{k}(w^{t}) - \nabla f(w^{t})\|^{2}) \\
    & + \frac{\eta}{2}(\eta L - 1) \mathbb{E}[\|\nabla f'_{k}(w^{t}) \|^{2}] - \frac{\eta}{2} \mathbb{E}[\|\nabla f(w^{t}) \|^{2}]. \\
\end{split}
\end{equation}
According to Cauchy-Schwarz inequality, \(\|\nabla f'_{k}(w^{t}) - \nabla f_{k}(w^{t}) + \nabla f_{k}(w^{t}) - \nabla f(w^{t})\|^{2} \leq (\|\nabla f'_{k}(w^{t}) - \nabla f_{k}(w^{t})\| + \|\nabla f_{k}(w^{t}) - \nabla f(w^{t})\|)^{2} \). Therefore, from Equation (\ref{eq:pt2}) we get:
\begin{equation}\label{eq:pt3}
\begin{split}
    & f(w^{t+1}) - f(w^{t}) \\
    & \leq \frac{\eta}{2} \mathbb{E}[(\|\nabla f'_{k}(w^{t}) - \nabla f_{k}(w^{t})\| + \|\nabla f_{k}(w^{t}) - \nabla f(w^{t})\|)^{2}] \\
    & + \frac{\eta}{2}(\eta L - 1) \mathbb{E}[\|\nabla f'_{k}(w^{t}) \|^{2}] - \frac{\eta}{2} \mathbb{E}[\|\nabla f(w^{t}) \|^{2}] \\
    & = \frac{\eta}{2}\mathbb{E}(\|\nabla f'_{k}(w^{t}) - \nabla f_{k}(w^{t})\|^{2} + \|\nabla f_{k}(w^{t}) - \nabla f(w^{t})\|^{2}) \\
    & + 2\mathbb{E}(\|\nabla f'_{k}(w^{t}) - \nabla f_{k}(w^{t})\|\times \|\nabla f_{k}(w^{t}) - \nabla f(w^{t})\|) \\
    & + \frac{\eta}{2}(\eta L - 1) \mathbb{E}[\|\nabla f'_{k}(w^{t}) \|^{2}] - \frac{\eta}{2} \mathbb{E}[\|\nabla f(w^{t}) \|^{2}] \\
    & \leq \frac{\eta}{2}(D^{2} + \gamma^{2} + 2D\gamma) + \frac{\eta}{2}(\eta L - 1) \mathbb{E}[\|\nabla f'_{k}(w^{t}) \|^{2}] \\
    & - \frac{\eta}{2} \mathbb{E}[\|\nabla f(w^{t}) \|^{2}].\\
\end{split}
\end{equation}
The last inequality in Equation (\ref{eq:pt3}) results from Assumption 2 and Assumption 3. \par
When the learning rate \(\eta > \frac{1}{L}\), we have \(\eta L- 1> 0\). In this case, we can upper bound \(\frac{\eta}{2}(\eta L - 1)\mathbb{E}[\|\nabla f'_{k}(w^{t}) \|^{2}]\) by upper bounding \(\mathbb{E}[\|\nabla f'_{k}(w^{t}) \|^{2}]\). \par
First, we bound \(\|\nabla f'_{k}(w^{t}) \|\). Based on Assumption 3 and the triangle inequality, we have:
\begin{equation}
    \|\nabla f'_{k}(w^{t})\| - \|\nabla f_{k}(w^{t})\| \leq \|\nabla f'_{k}(w^{t}) - \nabla f_{k}(w^{t}) \| \leq D.
\end{equation}
That is:
\begin{equation}\label{eq:pt4}
\begin{split}
    \|\nabla f'_{k}(w^{t})\|^{2} & \leq (\|\nabla f_{k}(w^{t})\|+D)^{2} \\
    & = \|\nabla f_{k}(w^{t})\| ^ {2} + D^{2} + 2D \|\nabla f_{k}(w^{t})\|.
\end{split}
\end{equation}
By taking the expectation on Equation (\ref{eq:pt4}), we get:
\begin{equation}\label{eq:pt5}
    \mathbb{E}[\|\nabla f'_{k}(w^{t})\|^{2}] \leq \mathbb{E}[\|\nabla f_{k}(w^{t})\|^{2}] + D^{2} + 2D \hspace{0.5mm} \mathbb{E}[\|\nabla f_{k}(w^{t})\|].
\end{equation}
Because of the triangle inequality, we have:
\begin{equation}\label{eq:boundf'}
\begin{split}
    \mathbb{E}[\|\nabla f_{k}(w^{t})\|] & = \mathbb{E}[\|\nabla f_{k}(w^{t}) - \nabla f (w^{t}) + \nabla f(w^{t})\|] \\
    & \leq \mathbb{E}[\|\nabla f_{k}(w^{t}) - \nabla f (w^{t})\|] + \mathbb{E}[\|\nabla f (w^{t})\|]  \\
    & \leq \mathbb{E}[\|\nabla f (w^{t})\|] + \gamma.  \\
\end{split}
\end{equation}

The last inequality in Equation (\ref{eq:boundf'}) holds because \(\mathbb{E}[\|\nabla f_{k}(w^{t}) - \nabla f (w^{t})\|] \leq \gamma\) as \((\mathbb{E}[\|\nabla f_{k}(w^{t}) - \nabla f (w^{t})\|])^{2} \leq \mathbb{E}[\|\nabla f_{k}(w^{t}) - \nabla f (w^{t})\|^{2}] \leq \gamma^{2}\) by Assumption 3. Moreover, by expanding Assumption 2, we have:
\begin{equation}\label{eq:pt6}
\begin{split}
    & \mathbb{E}[\|\nabla f_{k}(w^{t})\|^{2}] \\
    &  = \mathbb{E}[\|\nabla f_{k}(w^{t})- \nabla f(w^{t}) + \nabla f(w^{t})\|^{2}] \\
    & = \mathbb{E}[\|\nabla f(w^{t})\|^{2}] + \mathbb{E}[\|\nabla f_{k}(w^{t})- \nabla f(w^{t})\|^{2}] \\
    & +   2 \mathbb{E}(\langle\nabla f_{k}(w^{t})- \nabla f(w^{t}), \nabla f(w^{t})\rangle) \\
    & \leq \mathbb{E}[\|\nabla f(w^{t})\|^{2}] + \gamma^{2} + 2 \mathbb{E}(\langle\nabla f_{k}(w^{t})- \nabla f(w^{t}), \nabla f(w^{t})\rangle) \\
    & \leq  \mathbb{E}[\|\nabla f(w^{t})\|^{2}] + \gamma^{2} \\ 
    & + \mathbb{E}(\|\nabla f_{k}(w^{t})- \nabla f(w^{t})\|^{2} + \|\nabla f(w^{t})\|^{2}) \\
    & = 2 \mathbb{E}[\|\nabla f(w^{t})\|^{2}] + 2\gamma^{2} .\\
\end{split}
\end{equation}
By combining Equations (\ref{eq:pt5}), (\ref{eq:boundf'}), (\ref{eq:pt6}) altogether, we get:
\begin{equation}
\begin{split}
    & \mathbb{E}[\|\nabla f'_{k}(w^{t})\|^{2}] \\
    & \leq \mathbb{E}[\|\nabla f_{k}(w^{t})\|^{2}] + D^{2} + 2D \hspace{0.5mm} \mathbb{E}[\|\nabla f_{k}(w^{t})\|] \\
    & \leq 2 \mathbb{E}[\|\nabla f(w^{t})\|^{2}] + 2\gamma^{2} + D^{2} + 2D \hspace{0.5mm} \mathbb{E}[\|\nabla f_{k}(w^{t})\|] \\
    & \leq 2 \mathbb{E}[\|\nabla f(w^{t})\|^{2}] + 2\gamma^{2} + D^{2} + 2D (\mathbb{E}[\|\nabla f (w^{t})\|] + \gamma) \\
    & = 2 \mathbb{E}[\|\nabla f(w^{t})\|^{2}] + 2D \hspace{0.5mm} \mathbb{E}[\|\nabla f (w^{t})\|] + 2\gamma^{2} + D^{2} + 2D\gamma. \\
\end{split}
\end{equation}
Accordingly, we can rewrite Equation (\ref{eq:pt3}) as:
\begin{equation}\label{eq:pt3r1}
\begin{split}
    & f(w^{t+1}) - f(w^{t})  \\
    & \leq - \frac{\eta}{2} \mathbb{E}[\|\nabla f(w^{t}) \|^{2}] + \frac{\eta}{2}(D^{2} + \gamma^{2} + 2D\gamma) \\
    & +  \frac{\eta}{2}(\eta L - 1) \mathbb{E}[\|\nabla f'_{k}(w^{t}) \|^{2}] \\
    & \leq - \frac{\eta}{2} \mathbb{E}[\|\nabla f(w^{t}) \|^{2}] + \frac{\eta}{2}(D^{2} + \gamma^{2} + 2D\gamma) \\
    & + \frac{\eta}{2}(\eta L - 1)\times (2 \mathbb{E}[\|\nabla f(w^{t})\|^{2}] \\ 
    & + 2D \hspace{0.5mm} \mathbb{E}[\|\nabla f (w^{t})\|] + 2\gamma^{2} + D^{2} + 2D\gamma)  \\
    & = \frac{\eta}{2} (2\eta L -3) \mathbb{E}[\|\nabla f(w^{t})\|^{2}] \\ 
    & +\eta D (\eta L - 1) \mathbb{E}[\|\nabla f (w^{t})\|] \\ 
    & + \frac{\eta}{2} (2\eta L\gamma^{2} - \gamma^{2} + \eta L D^{2} + 2\eta L D \gamma).  \\
\end{split}
\end{equation}
When \(2\eta L-3 < 0\), i.e. \(\eta < \frac{3}{2L}\), we have \((2\eta L-3)\mathbb{E}[\|\nabla f(w^{t})\|^{2}] \leq (2\eta L-3) (\mathbb{E}[\|\nabla f(w^{t})\|)^{2}\). In this case, Equation (\ref{eq:pt3}) can be written as:
\begin{equation}\label{eq:pt3r2}
\begin{split}
    f(w^{t+1}) - f(w^{t}) & \leq \frac{\eta}{2} (2\eta L -3) (\mathbb{E}[\|\nabla f(w^{t})\|])^{2} \\ 
    & +\eta D (\eta L - 1) \mathbb{E}[\|\nabla f (w^{t})\|] \\ 
    & + \frac{\eta}{2} (2\eta L\gamma^{2} - \gamma^{2} + \eta L D^{2} + 2\eta L D \gamma).  \\
\end{split}
\end{equation}
which proves Theorem 1 successfully. Furthermore, if we take \(\mathbb{E}[\|\nabla f(w^{t})\|\) as a \textbf{variable}, \(f(w^{t+1}) - f(w^{t})\) is deemed to be upper bounded by a \textbf{polynomial function} of  \(\mathbb{E}[\|\nabla f(w^{t})\|\). In this case, we can naturally find \(\epsilon_{1} = \frac{-b-\sqrt{b^{2}-4ac}}{2a}\) by letting the polynomial function equal to zero, with \(a=2\eta L -3\), \(b=2D(\eta L-1)\) and \(c=2\eta L\gamma^{2} - \gamma^{2} + \eta L D^{2} + 2\eta L D \gamma\). \par
After calculation, we can get \(\epsilon_{1}=\)
\begin{equation}\label{epsilon1}
    \frac{D(\eta L-1) + \sqrt{\eta D^{2}L+8\eta L \gamma^{2} + 6\eta DL\gamma+D^{2}-3\gamma^{2}}}{3-2\eta L}
\end{equation}

Similarly, when the learning rate \(\eta \leq \frac{1}{L}\), we have \(\eta L- 1 \leq 0\). In this case, \(\frac{\eta}{2}(\eta L - 1)\mathbb{E}[\|\nabla f'_{k}(w^{t}) \|^{2}]\) is naturally upper bounded by zero, so that Equation (\ref{eq:pt3}) can be written as:
\begin{equation}
\begin{split}
f(w^{t+1}) - f(w^{t}) & \leq \frac{\eta}{2}(D^{2} + \gamma^{2} + 2D\gamma) - \frac{\eta}{2} \mathbb{E}[\|\nabla f(w^{t}) \|^{2}]. \\
\end{split}
\end{equation}
Which successfully proves Theorem 2. By letting \(\frac{\eta}{2}(D^{2} + \gamma^{2} + 2D\gamma) - \frac{\eta}{2} \mathbb{E}[\|\nabla f(w^{t}) \|^{2}]\) equal to zero we naturally get \(\epsilon_{2} = D + \gamma\). \par

For FedOLF with TOA, the above theorems remain valid. The only difference is that the boundary \(D\) in Assumption 3 is expected to become larger as TOA slightly increases the representation error. Subsequently, the critical points \(\epsilon_{1}\) and \(\epsilon_{2}\) also increase, 
resulting in an earlier halt in the decay of \(f\).

%% file: experiment.tex
\begin{table*}[ht]
    \centering
    \begin{tabular}{c|c|c|c|cc|cc}
    \toprule
    
        \multicolumn{2}{c|}{\textbf{Dataset}} & \textbf{EMNIST} & \textbf{CIFAR-10} & \multicolumn{2}{|c|}{\textbf{CIFAR-100}} & \multicolumn{2}{|c}{\textbf{CINIC-10}} \\
    
    \multicolumn{2}{c|}{Model} & CNN & AlexNet & ResNet20 & ResNet44 & ResNet20 & ResNet44 \\  
    \midrule
        \multicolumn{2}{c|}{Feddrop} & \(16.42\) & \(14.33\) & 6.05 & 6.15 & 9.71 & 10.81 \\
        \multicolumn{2}{c|}{FjORD}  & \(12.68\) & \(27.8\) & 11.14 & 9.09 & 22.22 & 11.26 \\
        \multicolumn{2}{c|}{HeteroFL} & 12.88 & 58.03 & 7.02 & 14.32 & 13.46 & 12.0 \\
        \multicolumn{2}{c|}{AdaptiveFL} & 34.4 & 49.42 & 18.37 & 25.13 & 14.25 & 15.44 \\
        \multicolumn{2}{c|}{ScaleFL} & 72.65 & 20.28 & 18.50 & 40.14 & 10.18 & 30.23 \\
        \multicolumn{2}{c|}{NeFL} & - & - & 39.55 & 40.29 & 32.97 & 31.68 \\
        \multicolumn{2}{c|}{DepthFL} & 83.0 & 10.52 & 5.05 & 39.88 & 10.31 & 33.44 \\
        \multicolumn{2}{c|}{CoCoFL} & \(81.98\) & \(53.92\) & 26.95 & 31.1 & 31.81 & 31.68\\
        \multicolumn{2}{c|}{SLT} & \(81.04\) & \(49.73\) & 45.80 & 39.60 & 21.63 & 36.20 \\
    \hline
        \multirow{3}{3.5em}{\textbf{FedOLF}} & no TOA & \textbf{84.98} & \textbf{66.98}& \textbf{48.49} & \textbf{44.12} & \textbf{40.66} & \textbf{37.33} \\
        & TOA(\(s=0.75\)) & - & \(63.7\) & \(40.49\) & 42.16 & 33.96 & 31.51 \\
        & TOA(\(s=0.5\)) & - & \(62.05\) & \(36.19\) & 38.29 & 33.42 & 28.42 \\
    \midrule
        \multicolumn{2}{c|}{FedAvg} & 85.04 & 68.41 & 51.11 & 52.13 & 40.80 & 39.88 \\
    \bottomrule
    \end{tabular}
    \caption{Comparison of the final test accuracy (in \%) for \(T=500\) iterations in the iid case. Note that for EMNIST where the number of frozen layers is at most one, FedOLF+TOA is not evaluated as TOA only works with at least two frozen layers.}
    \label{tab:acc_iid}
    \begin{tabular}{c|c|c|c|cc|cc}
    \toprule
    
        \multicolumn{2}{c|}{\textbf{Dataset}} & \textbf{EMNIST} & \textbf{CIFAR-10} & \multicolumn{2}{|c|}{\textbf{CIFAR-100}} & \multicolumn{2}{|c}{\textbf{CINIC-10}} \\
    
    \multicolumn{2}{c|}{Model} & CNN & AlexNet & ResNet20 & ResNet44 & ResNet20 & ResNet44 \\  
    \midrule
        \multicolumn{2}{c|}{Feddrop} & \(32.11\) & \(14.33\) & \(17.02\) & 6.2 & 9.87 & 10.31 \\
        \multicolumn{2}{c|}{FjORD}  & \(7.55\) & \(46.3\) & \(12.7\) & 14.68 & \(16.55\) & 20.08\\
        \multicolumn{2}{c|}{HeteroFL} & 17.4 & 54.79 & 12.32 & 12.96 & 10.69 & 10.03 \\
        \multicolumn{2}{c|}{AdaptiveFL} & 28.6 & 45.82 & 20.51 & 25.58 & 13.49 & 15.14 \\
        \multicolumn{2}{c|}{ScaleFL} & 56.64 & 13.96 & 13.73 & 35.51 & 9.74 & 33.59 \\
        \multicolumn{2}{c|}{NeFL} & - & - & 32.04 & 40.54 & 26.0 & 30.80 \\
        \multicolumn{2}{c|}{DepthFL} & 60.25 & 16.74 & 24.87 & 37.82 & 9.97 & 34.28 \\
        \multicolumn{2}{c|}{CoCoFL} & \(83.71\) & \(61.83\) & \(22.16\) & 27.56 & 25.66 & 26.67\\
        \multicolumn{2}{c|}{SLT} & \(60.72\) & \(30.47\) & \(25.04\) & 43.73 & 24.11 & 33.63 \\
    \hline
        \multirow{3}{3.5em}{\textbf{FedOLF}} & no TOA & \textbf{84.02} & \textbf{68.27}& \textbf{37.85} & \textbf{48.15} & \textbf{32.27} & \textbf{35.57}\\
        & TOA (\(s=0.75\)) & - & \(66.6\) & \(36.04\) & 40.72 & 31.85 & 32.52 \\
        & TOA (\(s=0.5\)) & - & \(63.12\) & \(24.93\) & 29.68 & 31.92 & 30.89 \\
    \midrule
        \multicolumn{2}{c|}{FedAvg} & 84.42 & 69.22 & 46.01 & 49.46 & 36.32 & 37.59 \\
    \bottomrule
    \end{tabular}
    \caption{Comparison of the final test accuracy (in \%) for \(T=500\) iterations in the non-iid case. For EMNIST where the number of frozen layers is at most one, FedOLF+TOA is not evaluated as TOA only works with at least two frozen layers.}
    \label{tab:acc}
\end{table*}

\section{Evaluation}\label{sec:exp}
We conduct extensive experiments to show the effectiveness of FedOLF with respect to accuracy, energy efficiency and memory efficiency. 
\subsection{Experiment Setup}

\textbf{Datasets and models:} We evaluate the performance of FedOLF on the Extended MNIST (EMNIST) \cite{emnist}, CIFAR-10 \cite{cifar}, CIFAR-100 \cite{cifar} and CINIC-10 \cite{cinic} datasets. For EMNIST, we adopt a convolutional neural network (CNN) consisting of two convolution layers and one fully-connected (FC) classifier \cite{fjord}. For CIFAR-10, we employ AlexNet \cite{alexnet} (five convolution layers + two FC layers). For CIFAR-100 and CINIC-10, we utilize ResNet20 and ResNet44 \cite{resnet}. \par

\textbf{State-of-the-art for comparison:} We compare FedOLF with the following representative methods for resource-constrained FL:
\begin{enumerate}
    \item \textbf{Federated Dropout (Feddrop)} \cite{randdrop} randomly prunes tensors in the global model and sends the remaining sub-model to clients for training.
    \item  \textbf{FjORD} \cite{fjord} prunes the rightmost tensors of the global model.
    \item \textbf{HeteroFL} \cite{heterofl} prunes the rightmost filters in convolution layers similar to FjORD, but keep the FC layers unchanged. 
    \item \textbf{AdaptiveFL} \cite{adaptivefl} implies an ordered dropout scheme same as in FjORD. In addition, it enables clients to maintain full shallow layers for better performance.  
    \item \textbf{DepthFL} \cite{depthfl} applies a top-first layer pruning method, and adds extra classifiers to clients with fewer layers to distill knowledge. 
    \item \textbf{ScaleFL} \cite{scalefl} utilizes the same layer pruning and knowledge distilllation strategy as in DepthFL. Additionally, it employs a leftmost dropout over neurons in unpruned layers to further enhance efficiency. 
    \item \textbf{CoCoFL} \cite{cocofl} let all clients store a full model locally and randomly freeze layers in training.
    \item \textbf{Successive Layer Training (SLT)} \cite{slt} mandates all clients to sequentially train each layer from bottom to top, while freezing the parameters of the remaining layers. 
    \item For CIFAR-100 and CINIC-10 with residual networks (ResNet20 and ResNet44), we also implement \textbf{NeFL} \cite{nefl}, which is an efficient FL framework tailored for residual model architectures. NeFL prunes the intermediate layers and intermittently trains the top and bottom layers by approximating the residual output with Taylor expansion. 
    \item Lastly, \textbf{FedAvg} \cite{fedavg} is also included as a standard benchmark to identify the theoretically highest accuracy without considering any resource constraints.
\end{enumerate}

\textbf{Parameter settings and system implementation:} The experiment runs on a virtual network consisting of \(K=100\) clients operating on a desktop computer with one NVIDIA GeForce GTX 1650 GPU. The number of participants per round is \(|C_{t}|=10\) following the settings in \cite{fjord, randdrop}. The maximum global iteration is set to \(T=500\) and the local training epoch is \(E=5\) for all clients \cite{fjord, hermes}. The learning rate is set to \(\eta=0.0001\) for EMNIST, \(\eta=0.001\) for CIFAR-10, and \(\eta=0.01\) for CIFAR-100 and CINIC-10 \cite{nofear}. The batch size is set to 16 for EMNIST and 128 for the remaining datasets \cite{hermes, fjord}. The experiment is implemented with PyTorch 2.0.0 and Flower 1.4.0 \cite{flwr}. \par

\textbf{Client data heterogeneity:} We evaluate FedOLF in both iid and non-iid environments. For the iid case, data are allocated to clients uniformly. For the non-iid case, we follow \cite{nofear} and allocate data to clients based on an extreme Dirichlet distribution with parameter 0.1. \par

\textbf{Client system heterogeneity:} To emulate system heterogeneity, we randomly divide all clients into \(c\) uniform clusters that represent \(c\) different degrees of device capability and resource constraints, as per \cite{fjord,heterofl,depthfl,cocofl}. Specifically, for EMNIST that runs on two-layer CNNs, we naturally set \(c=2\), representing zero and one frozen/pruned layer for layer-wise methods, including FedOLF, CoCoFL, DepthFL and NeFL. For the other datasets and models, we set \(c=5\) following \cite{fjord}, with \(\{4,3,2,1,0\}\) frozen/pruned layers (for AlexNet) and residual blocks (for ResNet20 and ResNet44) following \cite{depthfl}. \par

For dropout-based methods \cite{randdrop, fjord, heterofl, adaptivefl, scalefl}, we follow \cite{fjord, heterofl} and assign uniform sub-model ratios (i.e., the percentage of left neurons per layer) \(\{\frac{1}{c},......,\frac{c}{c}\}\) to the \(c\) clusters. Therefore, for CNN on EMNIST,  the sub-model ratios are 0.5 and 1.0 for the two clusters. For other datasets with \(c=5\), the sub-model ratios are \(\{0.2,0.4,0.6,0.8,1.0\}\). \par
For SLT \cite{slt} that conducts universal successive training among all clients, the scaling factor for the partial training procedure is initially set to 0.5 and gradually increases to 1.0.

\textbf{Metrics:} We evaluate the performance of FedOLF in multiple aspects:
\begin{itemize}
    \item We evaluate the \textbf{accuracy} in both iid and non-iid cases. All methods have run for three independent trials, and their mean performance is recorded.
    \item We evaluate the \textbf{total computation and communication costs} in terms of energy consumption following \cite{flrce, hermes}. Concretely, we first measure the power rate of our experimental device using a power monitor\footnote{ https://www.amazon.com.au/Electricity-Monitor-PIOGHAX-Overload-Protection/dp/B09SFSB66M.}, then derive the overall computation/communication cost in terms of energy, which is equal to the product of the power rate and the total computation/communication time of all clients. 
    \item Combining accuracy and energy consumption, we derive the \textbf{energy efficiency} by observing the highest achievable accuracy with the same amount of energy consumption.
    \item We evaluate both the practical and theoretical \textbf{memory consumptions.} The practical memory consumption is measured using the {\fontfamily{qcr}\selectfont TORCH.CUDA.MAX\underline{\hspace{1mm}}MEMORY\underline{\hspace{1mm}}ALLOCATED} function \cite{cudamemory}, and the theoretical memory consumption is the summation of weights, gradients and activations stored in all parameters \cite{slt}. 
\end{itemize}

\subsection{Experiment Results}
\textbf{Accuracy:} Tables \ref{tab:acc_iid} and \ref{tab:acc} respectively show the accuracy comparison in the iid and non-iid cases. As shown in Tables \ref{tab:acc_iid} and \ref{tab:acc}, FedOLF achieves the highest final accuracy among all methods on all datasets, which demonstrates the strength of FedOLF in preserving accuracy on resource-constrained devices. By looking through all methods, we find that dropout (Feddrop, FjORD, AdaptiveFL, ScaleFL) performs poorly with non-iid data, as training a sub-model cannot extract sufficient knowledge from the local dataset to construct an accurate global model \cite{cocofl}. Furthermore, sub-models with inconsistent architecture often learn divergent parameter updates during training, and aggregating these updates together will inevitably compromise the global model's performance \cite{prunefl, depthfl}. Although existing layer-freezing approaches (CoCoFL, SLT) improve accuracy by maintaining the full model architecture on all clients, they still lag behind FedOLF in accuracy. The possible reason is that FedOLF incurs fewer training errors by sharing the well-generalized low-level layers among clients, as discussed in Section \ref{subsec:fedolf}. While in CoCoFL or SLT, larger training errors occur when clients share the diverse frozen top layers. \par

\begin{figure}[ht]
    \centering
    \begin{subfigure}[t]{0.24\textwidth}
        \centering
        \includegraphics[width=\textwidth, height=0.15\textheight]{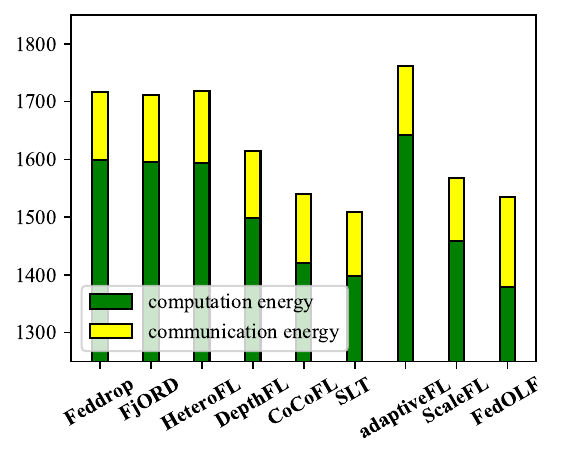}
        \caption{EMNIST.}
    \end{subfigure}
    \begin{subfigure}[t]{0.24\textwidth}
        \centering
        \includegraphics[width=\textwidth, height=0.151\textheight]{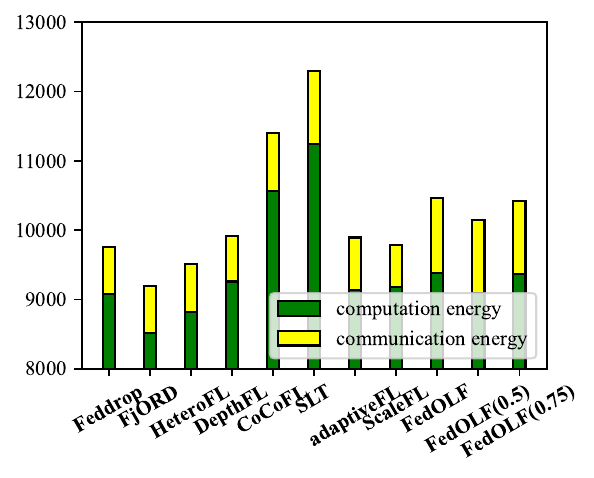}
        \caption{CIFAR-10.}
    \end{subfigure}
    \begin{subfigure}[t]{0.24\textwidth}
        \centering
        \includegraphics[width=\textwidth]{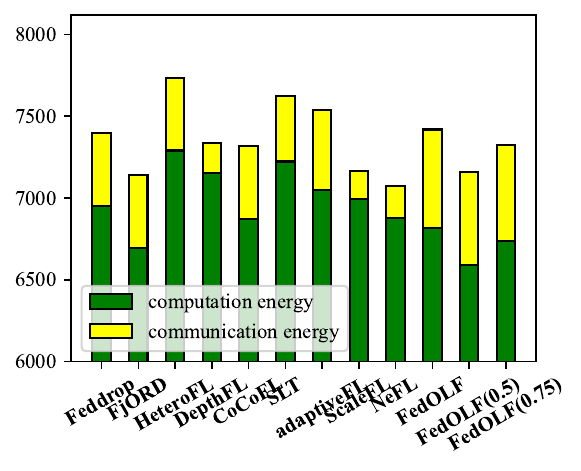}
        \caption{CIFAR-100 on ResNet20.}
    \end{subfigure}
    \begin{subfigure}[t]{0.24\textwidth}
        \centering
        \includegraphics[width=\textwidth,
        ]{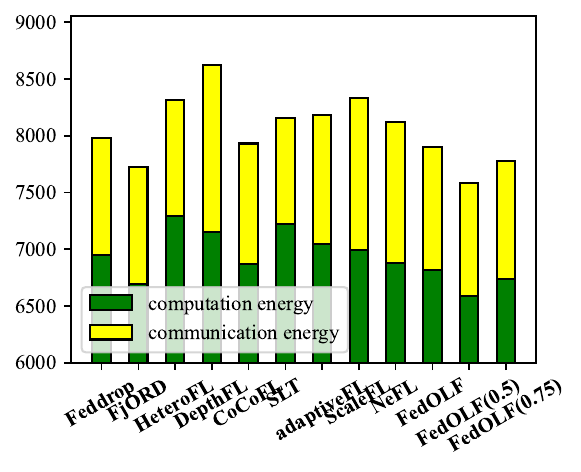}
        \caption{CIFAR-100 on ResNet44.}
    \end{subfigure}
    \begin{subfigure}[t]{0.24\textwidth}
        \centering
        \includegraphics[width=\textwidth,
        ]{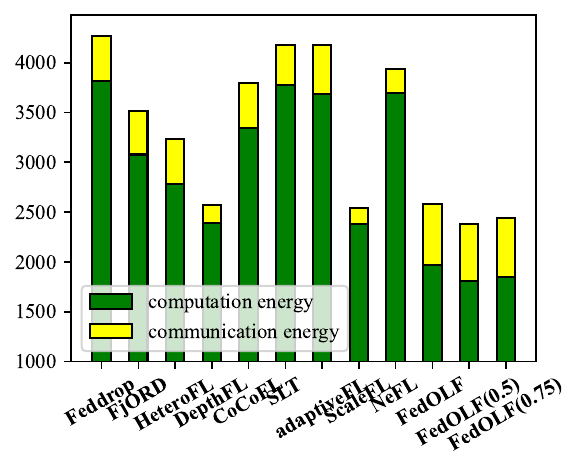}
        \caption{CINIC-10 on ResNet20.}
    \end{subfigure}
    \begin{subfigure}[t]{0.24\textwidth}
        \centering
        \includegraphics[width=\textwidth,
        ]{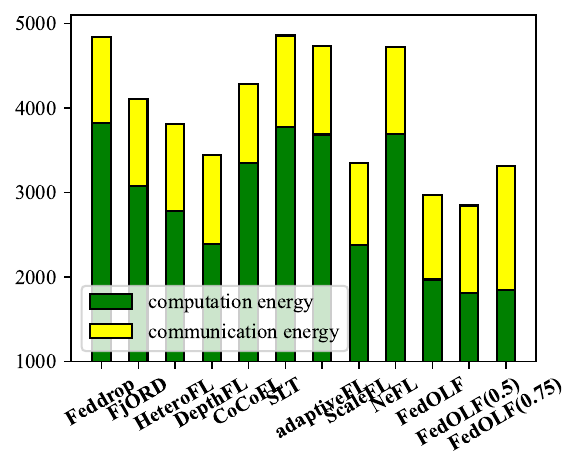}
        \caption{CINIC-10 on ResNet44.}
    \end{subfigure}
    \caption{An overview of the overall energy consumption (kJ) of all clients, including the computation energy for local training (green) and the communication energy for global communication (yellow). }
    \label{fig:overall_cost}
\end{figure}

Although FedOLF incurs a marginal accuracy loss compared with FedAvg in most scenarios, the accuracy degradation deteriorates for CIFAR-100 with ResNet20 on non-iid data (see Table \ref{tab:acc}). We analyze that this is due to the following possible reasons, which we plan to address in future work:
\begin{itemize}
    \item \textbf{Representation diverse:} For a large degree of non-iid data distribution, the low-level representation output is likely to diverge among clients \cite{nofear}. Simply sharing the bottom layers without considering the representation diversity no longer guarantees performance.
    \item \textbf{Mismatch between data importance and system capacity:} In the non-iid environment, some clients are more important than others. That is, their local data distributions resemble the global distribution \cite{flrce}. If these clients obtain weak devices with more frozen layers, the learning efficiency of FL will be impeded.
    \item \textbf{Vulnerability of small models:} For complex tasks, every single layer is crucial for a small model to maintain performance. Subsequently, small models are more vulnerable to training loss caused by layer freezing. In comparison, the accuracy loss of ResNet44 is much smaller as shown in Table \ref{tab:acc}.
\end{itemize}

\begin{figure}[htbp]
    \centering
    \begin{subfigure}[t]{0.5\textwidth}
        \centering
        \includegraphics[width=\textwidth]{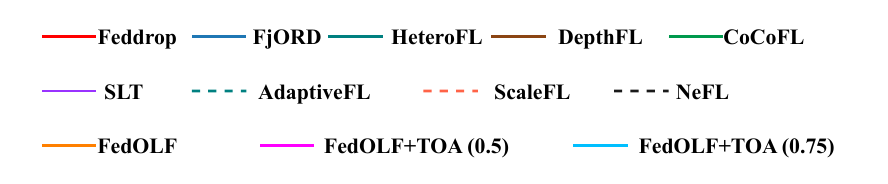}
    \end{subfigure}
    \begin{subfigure}[t]{0.24\textwidth}
        \centering
        \includegraphics[width=\textwidth]{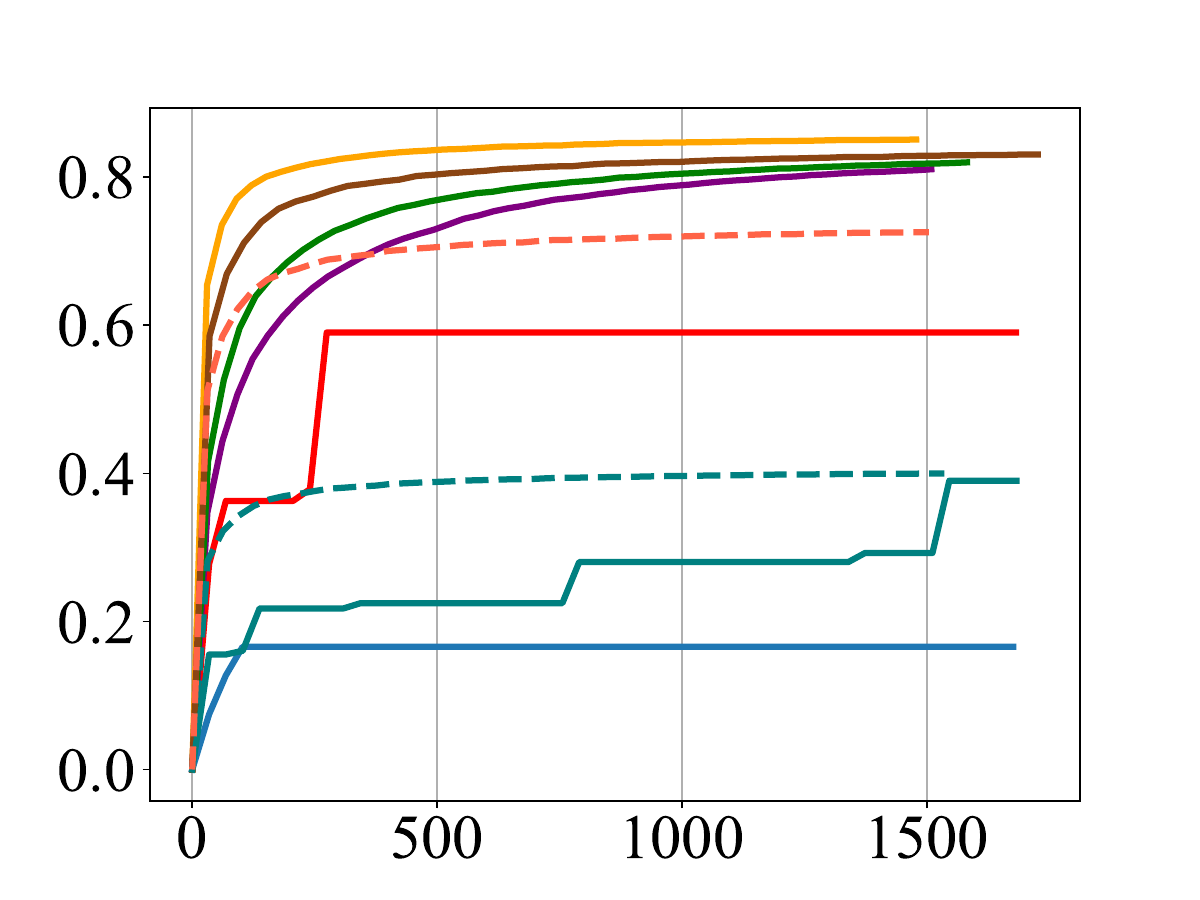}
        {\scriptsize Energy consumption (kJ)}
        \caption{EMNIST.}
    \end{subfigure}
    \hfill
    \begin{subfigure}[t]{0.24\textwidth}
        \centering
        \includegraphics[width=\textwidth,
        ]{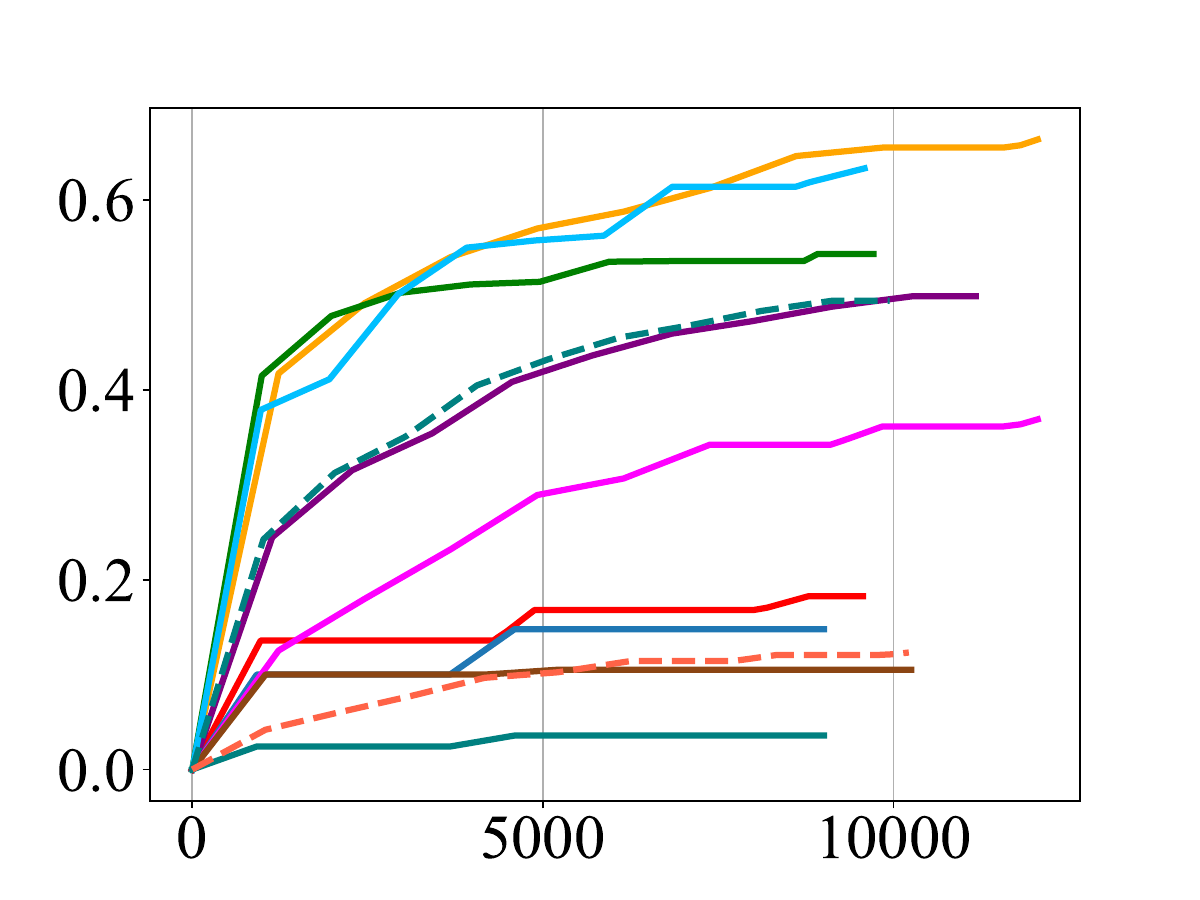}
        {\scriptsize Energy consumption (kJ)}
        \caption{CIFAR-10.}
    \end{subfigure}
    \hfill
    \begin{subfigure}[t]{0.24\textwidth}
        \centering
        \includegraphics[width=\textwidth,
        ]{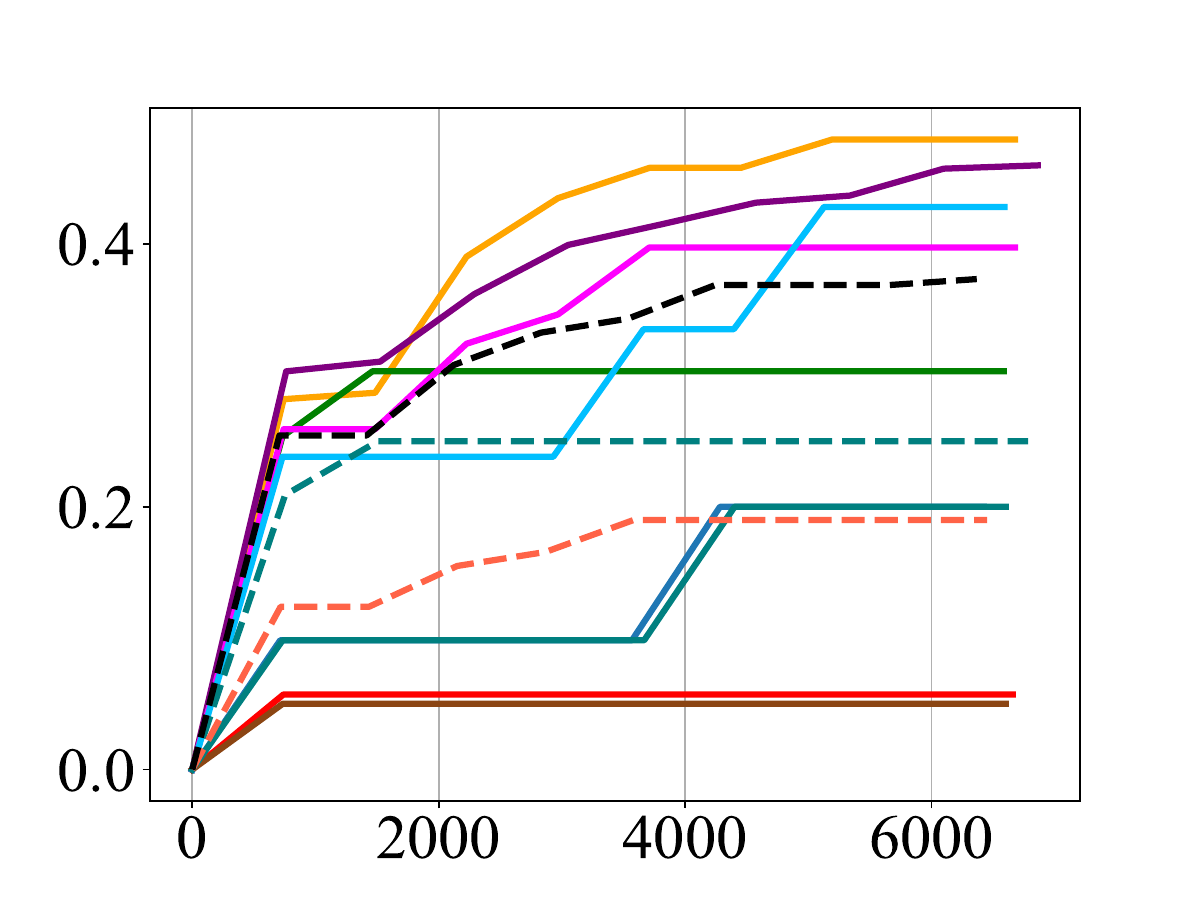}
        {\scriptsize Energy consumption (kJ)}
        \caption{CIFAR-100 on ResNet 20.}
    \end{subfigure}
    \hfill
    \begin{subfigure}[t]{0.24\textwidth}
        \centering
        \includegraphics[width=\textwidth,
        ]{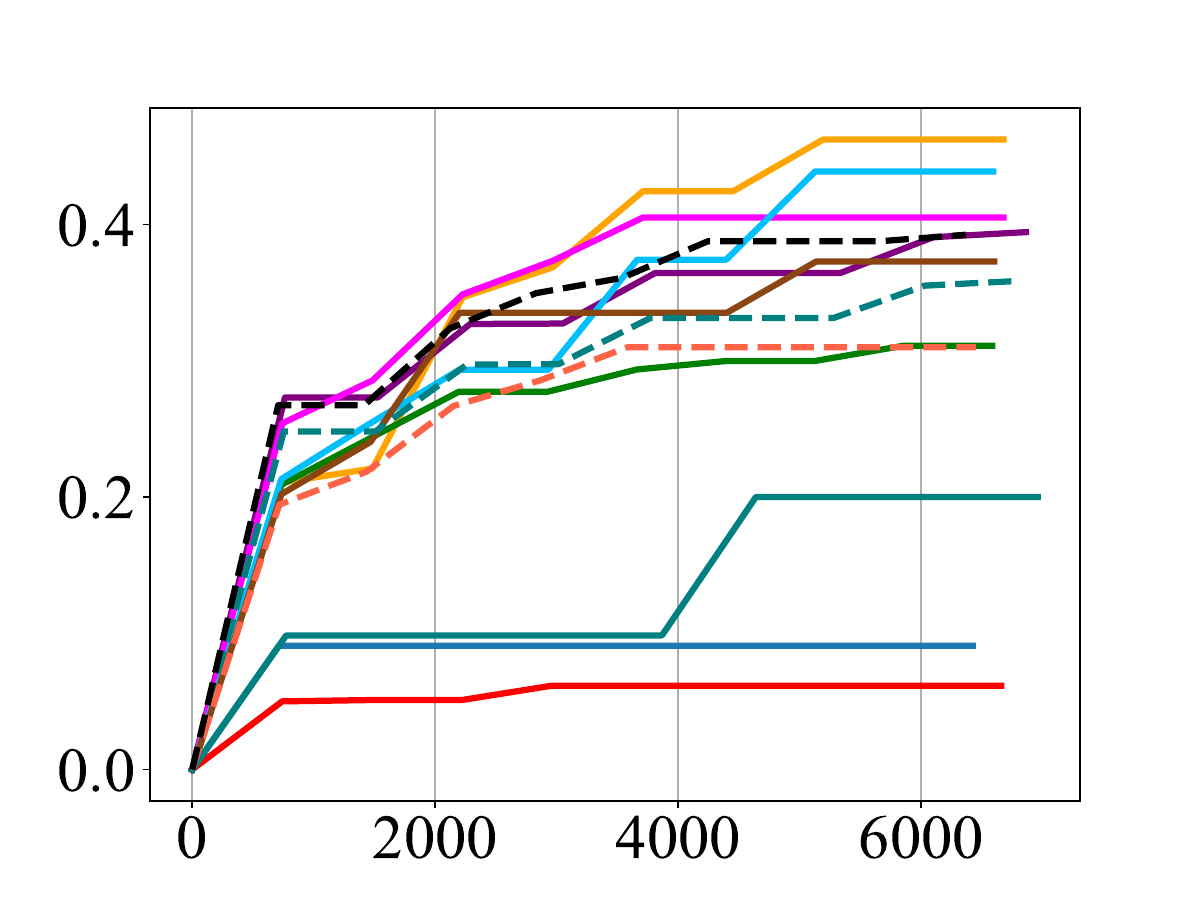}
        {\scriptsize Energy consumption (kJ)}
        \caption{CIFAR-100 on ResNet 44.}
    \end{subfigure}
    \begin{subfigure}[t]{0.24\textwidth}
        \centering
        \includegraphics[width=\textwidth,
        ]{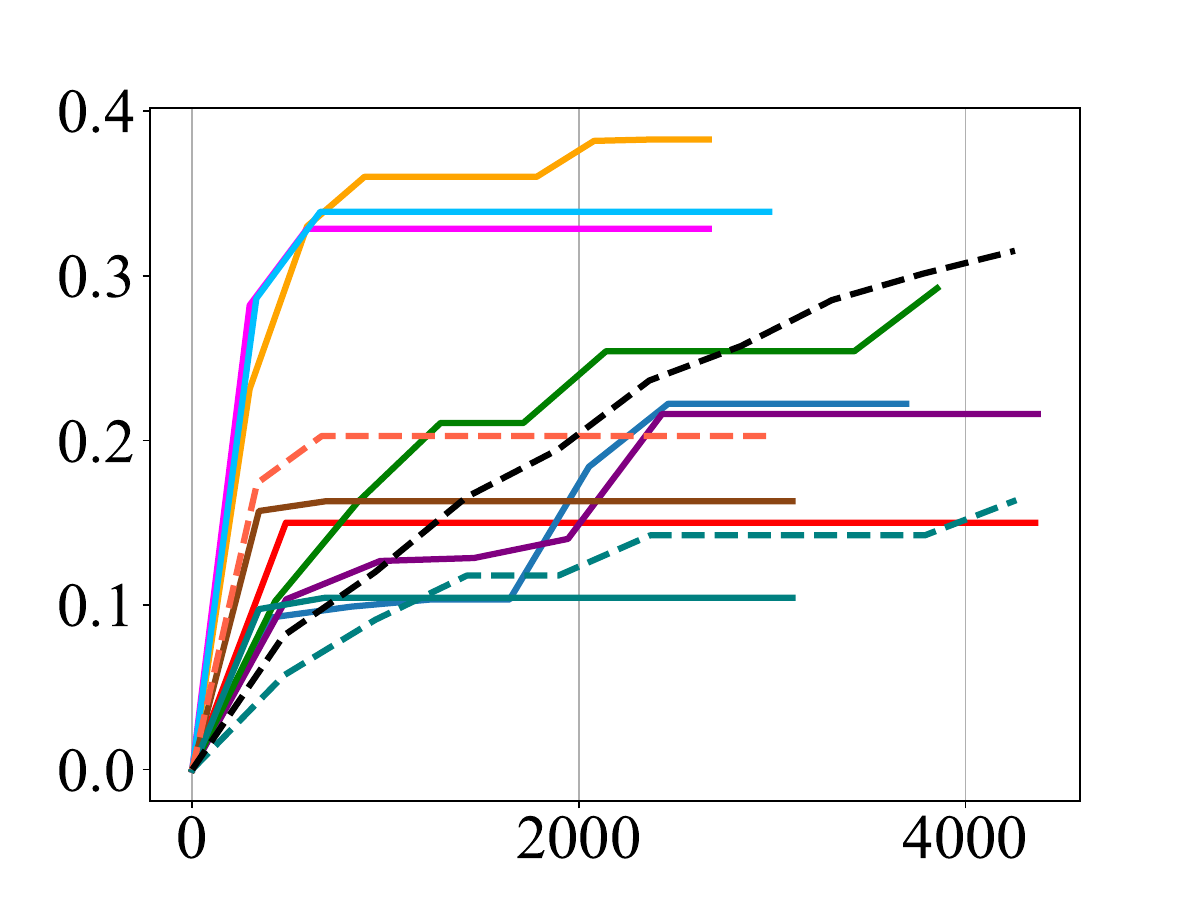}
        {\scriptsize Energy consumption (kJ)}
        \caption{CINIC-10 on ResNet 20.}
    \end{subfigure}
    \begin{subfigure}[t]{0.24\textwidth}
        \centering
        \includegraphics[width=\textwidth,
        ]{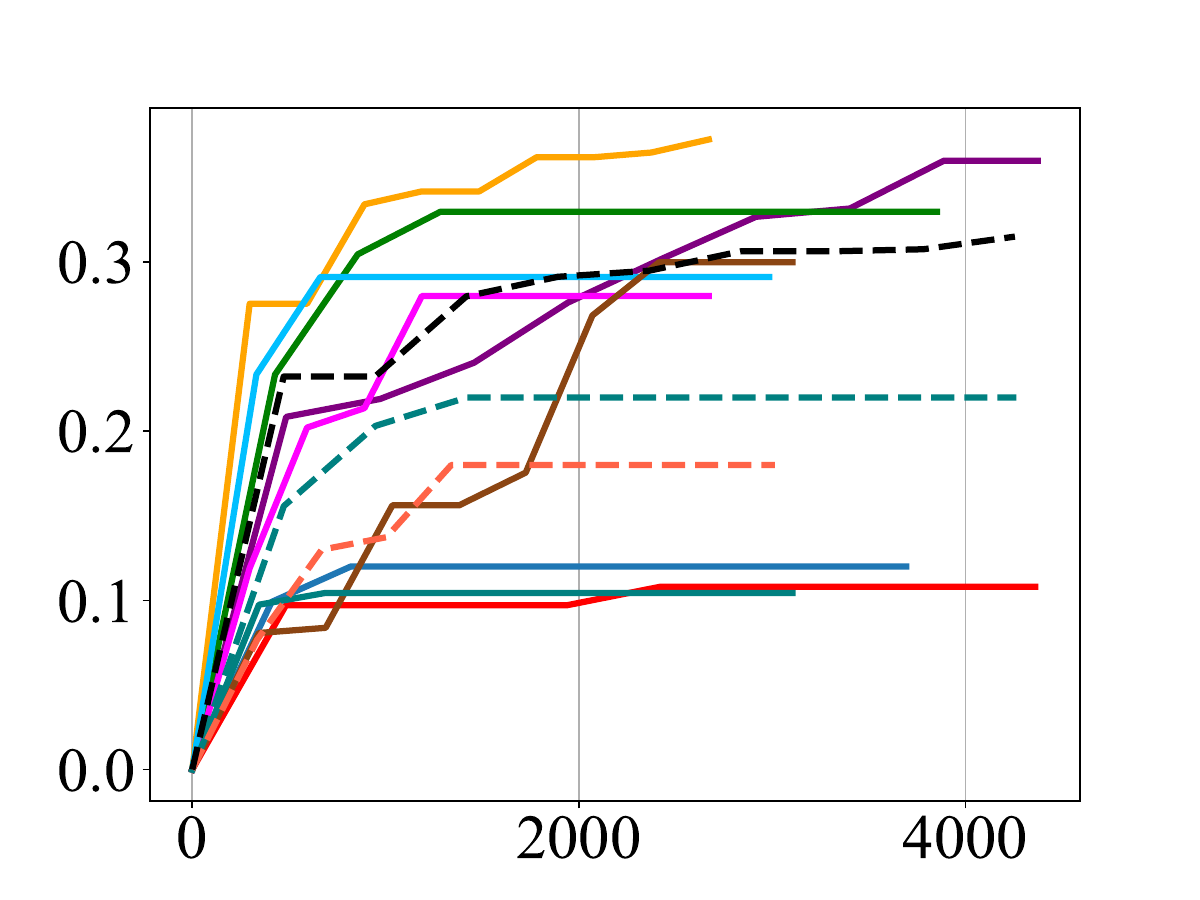}
        {\scriptsize Energy consumption (kJ)}
        \caption{CINIC-10 on ResNet 44.}
    \end{subfigure}
    \caption{The curves of top-1 accuracy (in the iid case) vs. energy consumption (kJ).}
    \label{fig:overall_efficiency_iid}
\end{figure}

\textbf{Computation, communication costs, and energy efficiency:} Figure \ref{fig:overall_cost} summarizes the total energy consumption in the non-iid case (the costs in the iid case are almost equal). This encompasses the computation energy for training and the communication energy for parameter transmission. From Figure \ref{fig:overall_cost}, FedOLF does not consume much additional energy compared with others. Even in the worst case (CIFAR-10), FedOLF only consumes \(13\%\) more energy than the least of the baselines (FjORD). This is a reasonable trade-off considering the immense accuracy improvement (see Tables \ref{tab:acc_iid} and \ref{tab:acc}). In some cases, FedOLF even reduces the overall energy consumption (e.g. CINIC-10). This can be attributed to the enhanced effectiveness of ordered layer freezing in reducing computation overhead for deeper neural networks. \par

\begin{figure}[htbp]
    \centering
    \begin{subfigure}[t]{0.5\textwidth}
        \centering
        \includegraphics[width=\textwidth]{image/efficiency_legend_singlecolumn}
    \end{subfigure}
    \begin{subfigure}[t]{0.24\textwidth}
        \centering
        \includegraphics[width=\textwidth]{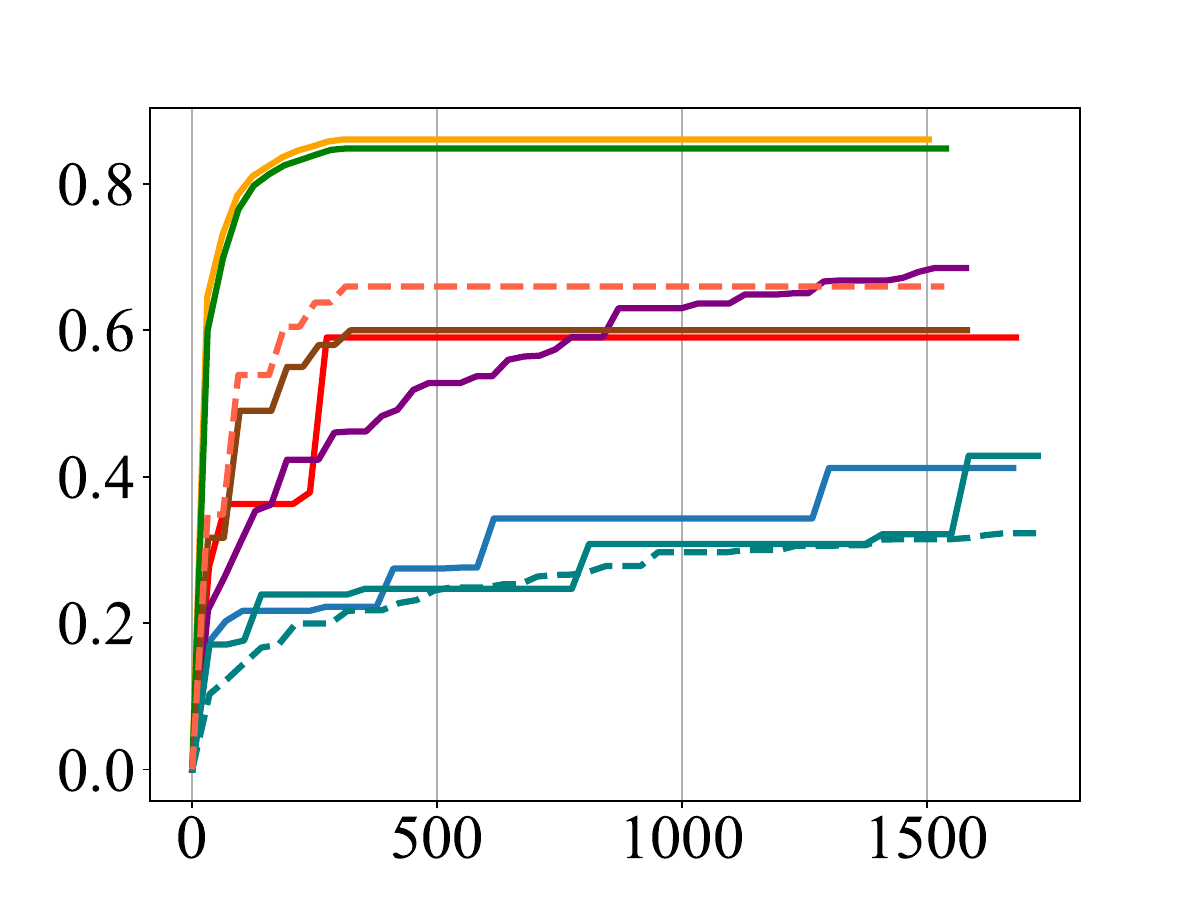}
        {\scriptsize Energy consumption (kJ)}
        \caption{EMNIST.}
    \end{subfigure}
    \hfill
    \begin{subfigure}[t]{0.24\textwidth}
        \centering
        \includegraphics[width=\textwidth,
        ]{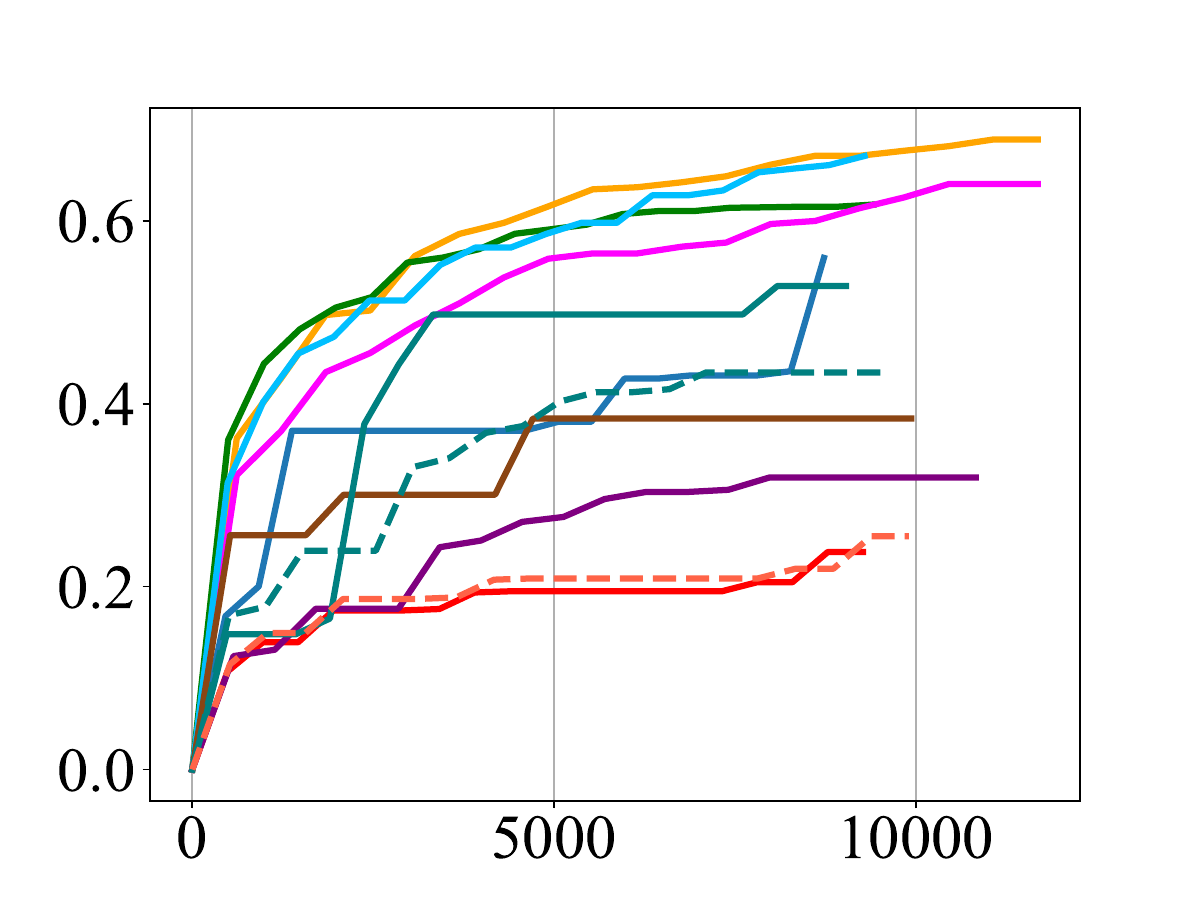}
        {\scriptsize Energy consumption (kJ)}
        \caption{CIFAR-10.}
    \end{subfigure}
    \hfill
    \begin{subfigure}[t]{0.24\textwidth}
        \centering
        \includegraphics[width=\textwidth,
        ]{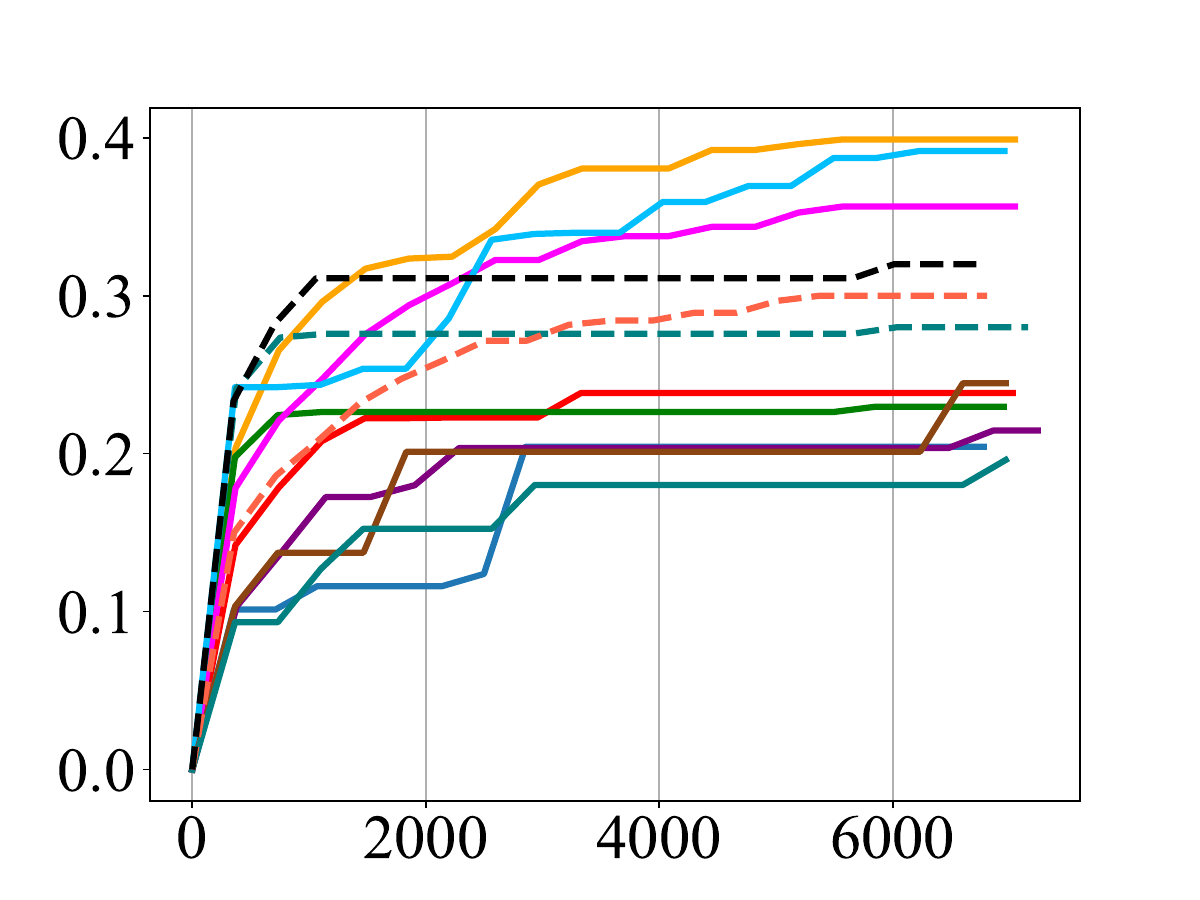}
        {\scriptsize Energy consumption (kJ)}
        \caption{CIFAR-100 on ResNet 20.}
    \end{subfigure}
    \hfill
    \begin{subfigure}[t]{0.24\textwidth}
        \centering
        \includegraphics[width=\textwidth,
        ]{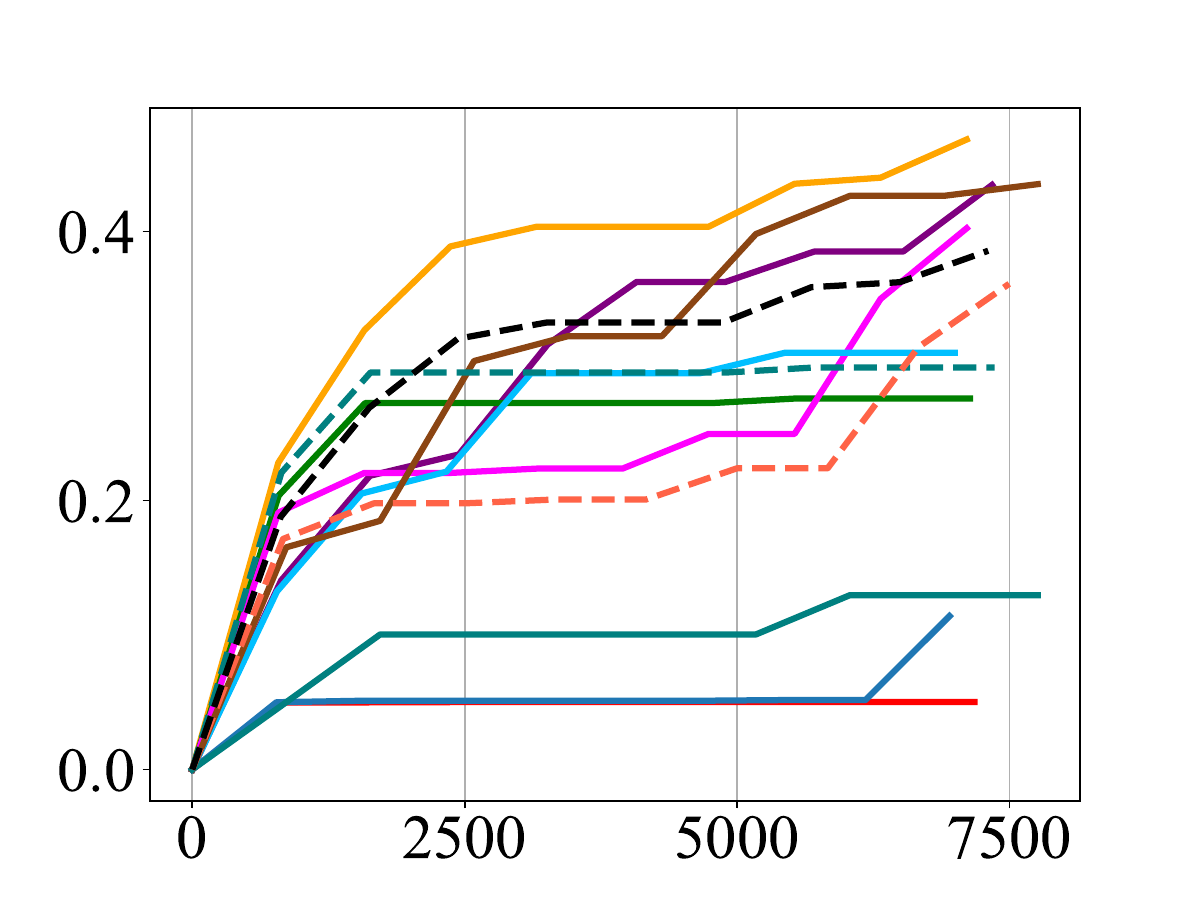}
        {\scriptsize Energy consumption (kJ)}
        \caption{CIFAR-100 on ResNet 44.}
    \end{subfigure}
    \begin{subfigure}[t]{0.24\textwidth}
        \centering
        \includegraphics[width=\textwidth,
        ]{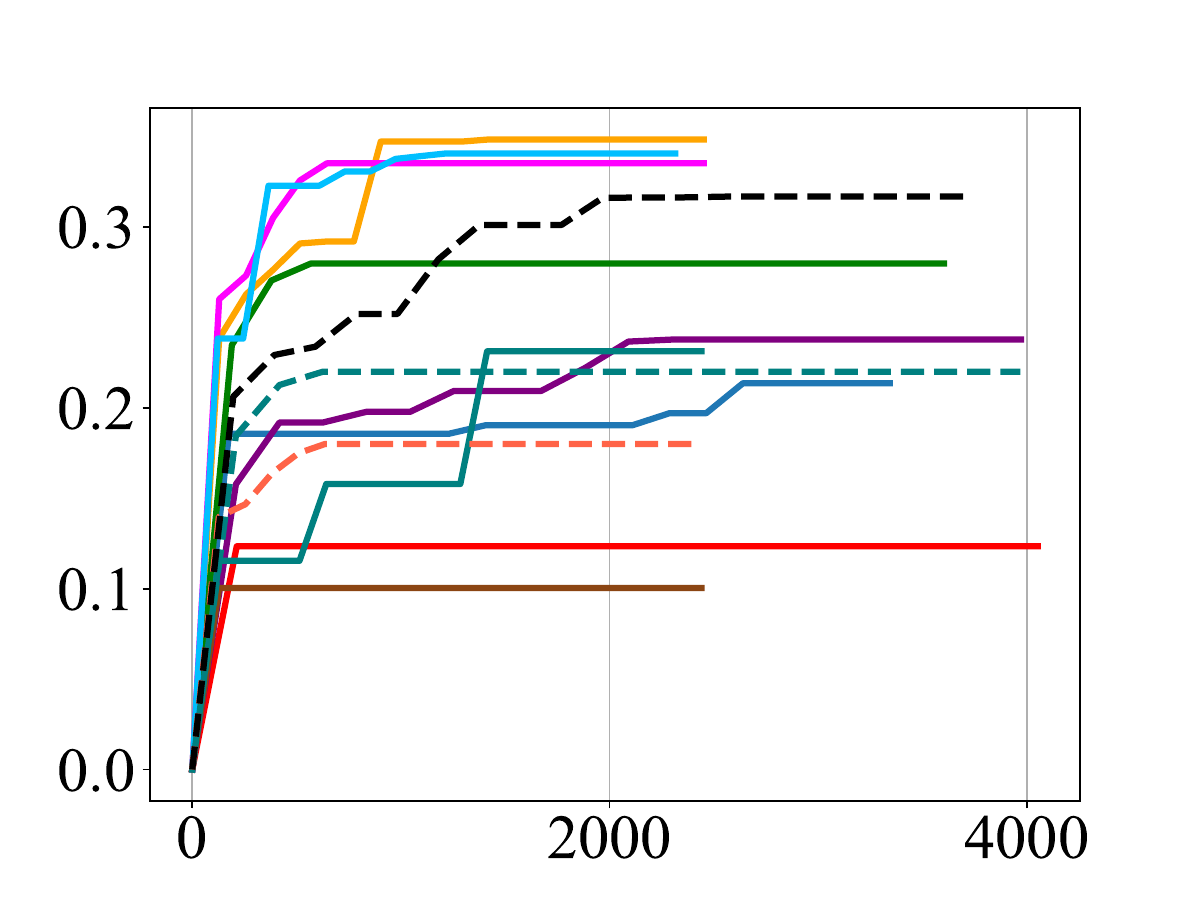}
        {\scriptsize Energy consumption (kJ)}
        \caption{CINIC-10 on ResNet 20.}
    \end{subfigure}
    \begin{subfigure}[t]{0.24\textwidth}
        \centering
        \includegraphics[width=\textwidth,
        ]{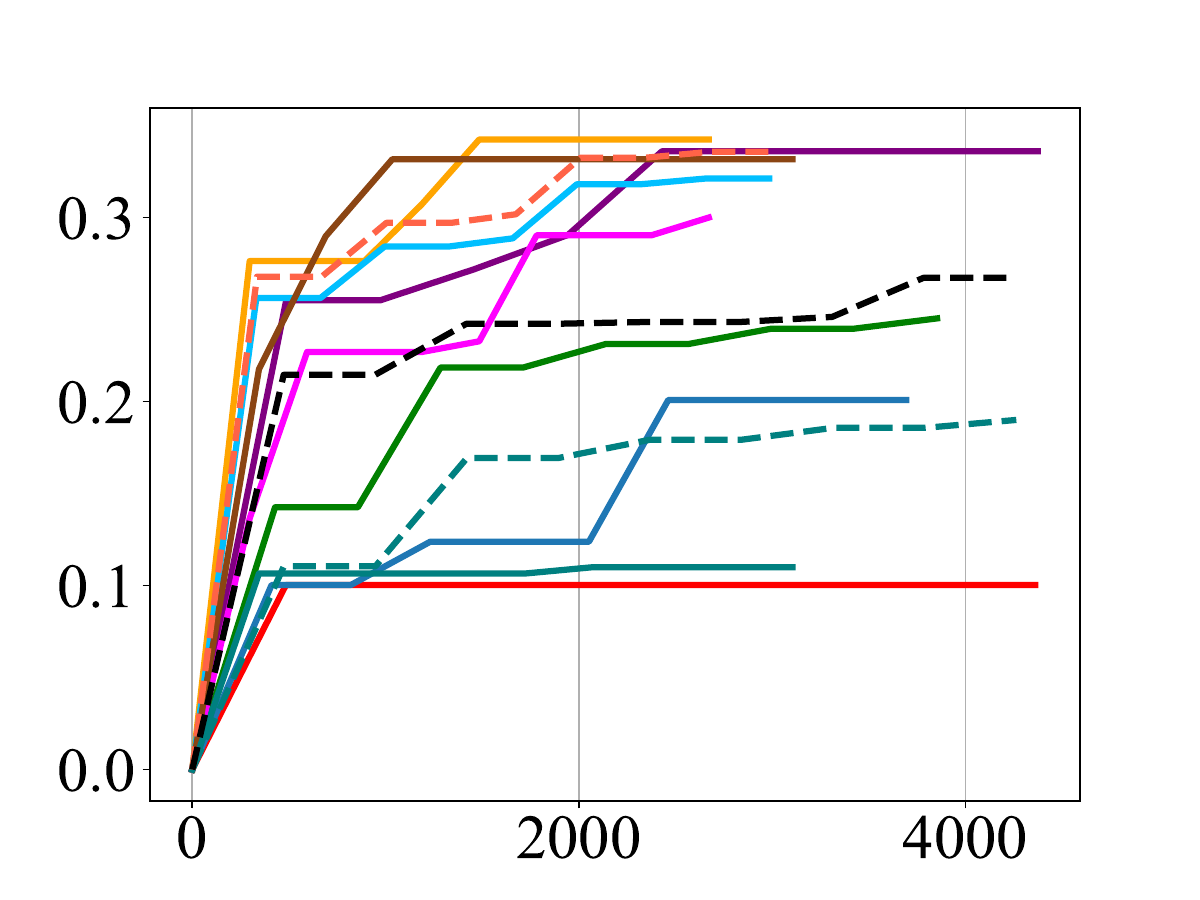}
        {\scriptsize Energy consumption (kJ)}
        \caption{CINIC-10 on ResNet 44.}
    \end{subfigure}
    \caption{The curves of top-1 accuracy (in the non-iid case) vs. energy consumption (kJ).}
    \label{fig:overall_efficiency}
\end{figure}

For fairness, we derive the overall energy efficiency of all methods by plotting the highest accuracy achieved versus the corresponding energy consumption. As Figures \ref{fig:overall_efficiency_iid} and \ref{fig:overall_efficiency} show, for both iid and non-iid data distributions, the curve of FedOLF surpasses others in most cases. This implies that, although FedOLF does not usually obtain the least overall energy consumption, it usually achieves the highest accuracy given the same amount of energy expenditure. In this case, for resource-constrained systems that cannot afford the full FL iterations due to an insufficient energy budget, FedOLF remains the optimal framework to maximize the possibly achievable accuracy. \par 

In addition, although some methods can quickly achieve appropriate accuracy, they cannot maintain that accuracy due to the lack of a convergence guarantee. For example, although Feddrop can quickly obtain a high accuracy (\(\approx60\%\)) on the MNIST dataset (see Figure \ref{fig:overall_efficiency_iid} (a)), the ultimate accuracy of Feddrop is low (\(16.42\%\)) as shown in Table \ref{tab:acc_iid}. Comparatively, the ultimate accuracies of FedOLF in Figures \ref{fig:overall_efficiency_iid} and \ref{fig:overall_efficiency} are very close to the final test accuracies in Tables \ref{tab:acc_iid} and \ref{tab:acc}. This demonstrates the convergence property that enables FedOLF to maintain the achieved accuracy.

\begin{figure}[htbp]
    \centering
    \begin{subfigure}[t]{0.5\textwidth}
        \centering
        \includegraphics[width=\textwidth]{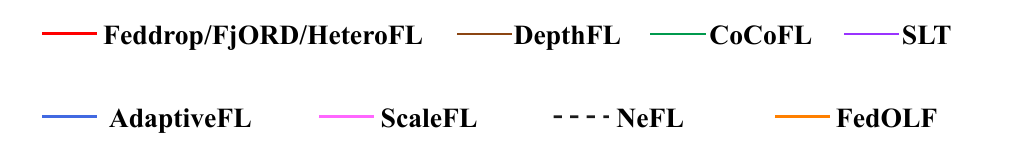}
    \end{subfigure}
    \begin{subfigure}[t]{0.24\textwidth}
        \centering
        CNN
        \includegraphics[width=\textwidth]{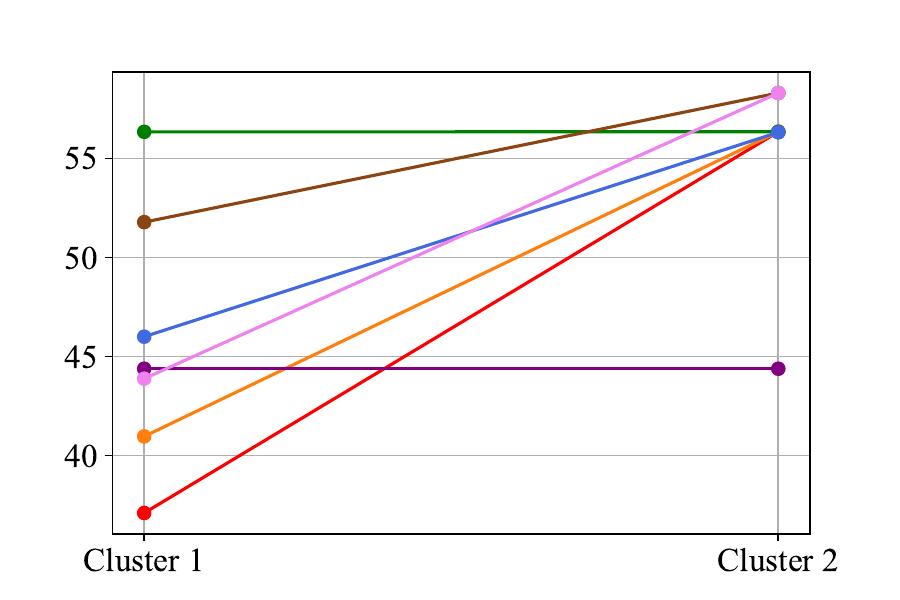}
    \end{subfigure}
    \begin{subfigure}[t]{0.24\textwidth}
        \centering
        AlexNet
        \includegraphics[width=\textwidth,
        ]{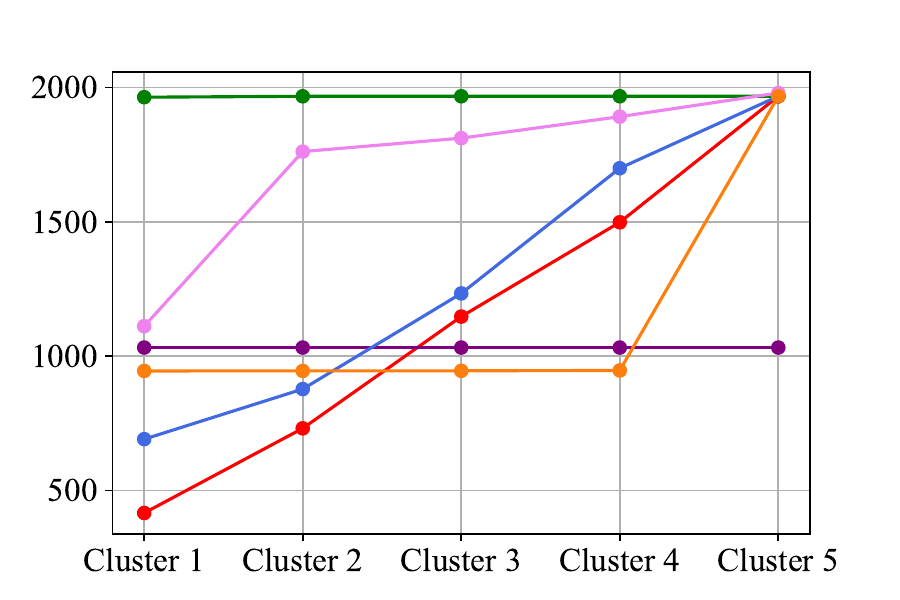}
    \end{subfigure}
    \begin{subfigure}[t]{0.24\textwidth}
        \centering
        ResNet20
        \includegraphics[width=\textwidth,
        ]{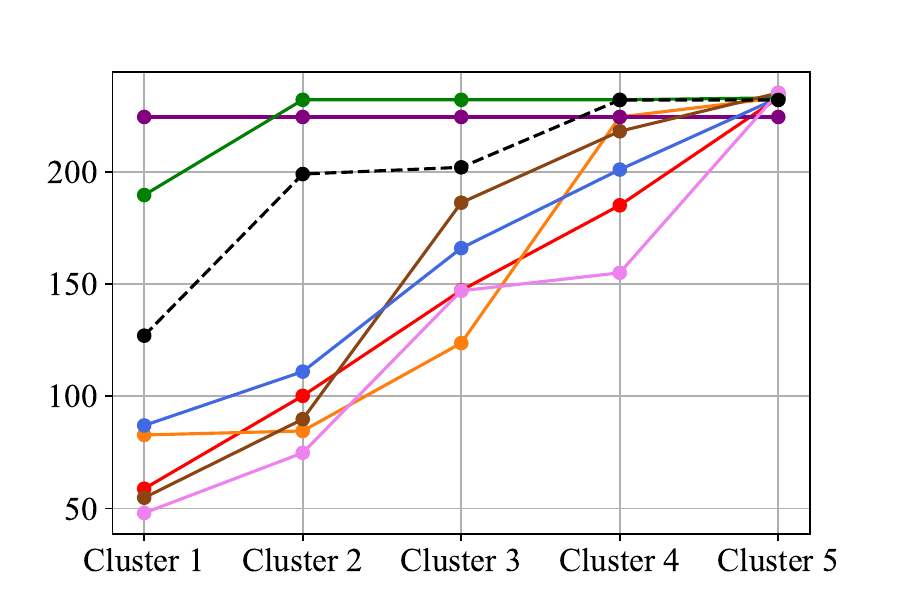}
    \end{subfigure}
    \begin{subfigure}[t]{0.24\textwidth}
        \centering
        ResNet44
        \includegraphics[width=\textwidth,
        ]{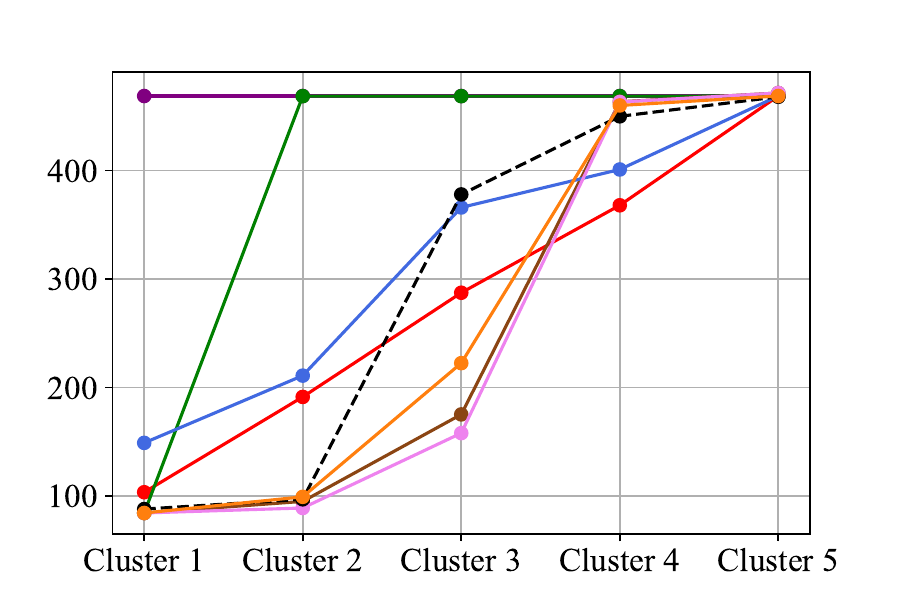}
    \end{subfigure}
    \caption{The actual memory consumption (MB) among clients.}\label{fig:act_mem}
\end{figure}
\begin{figure}[htbp]
    \centering
    \begin{subfigure}[t]{0.5\textwidth}
        \centering
        \includegraphics[width=\textwidth]{image/fedolf_memory_legend}
    \end{subfigure}
    \begin{subfigure}[t]{0.24\textwidth}
        \centering
        CNN
        \includegraphics[width=\textwidth]{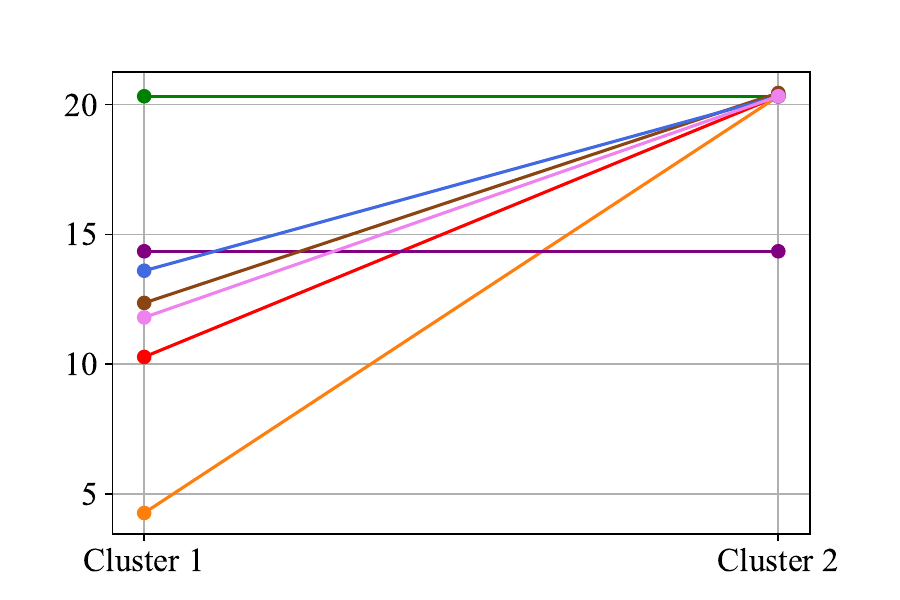}
    \end{subfigure}
    \begin{subfigure}[t]{0.24\textwidth}
        \centering
        AlexNet
        \includegraphics[width=\textwidth,
        ]{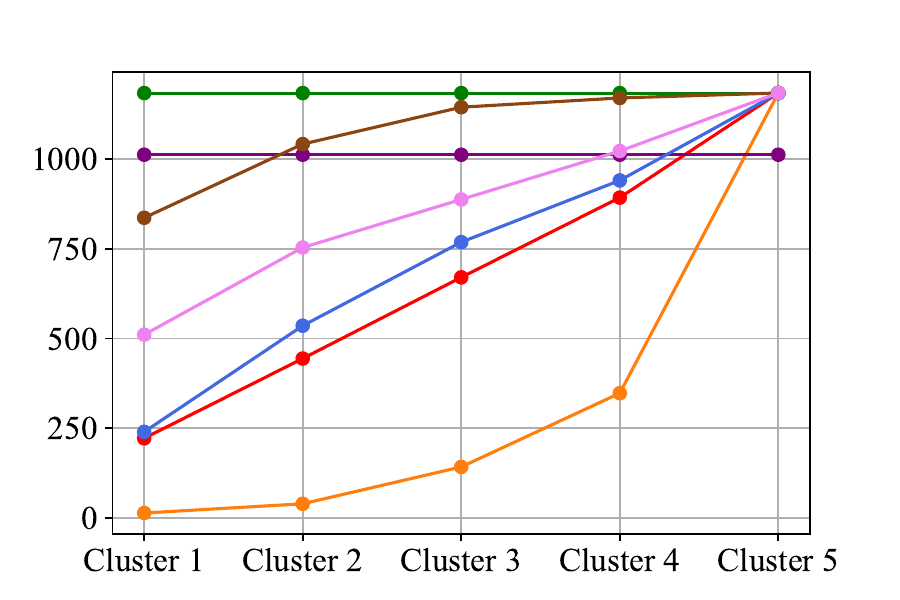}
    \end{subfigure}
    \begin{subfigure}[t]{0.24\textwidth}
        \centering
        ResNet20
        \includegraphics[width=\textwidth,
        ]{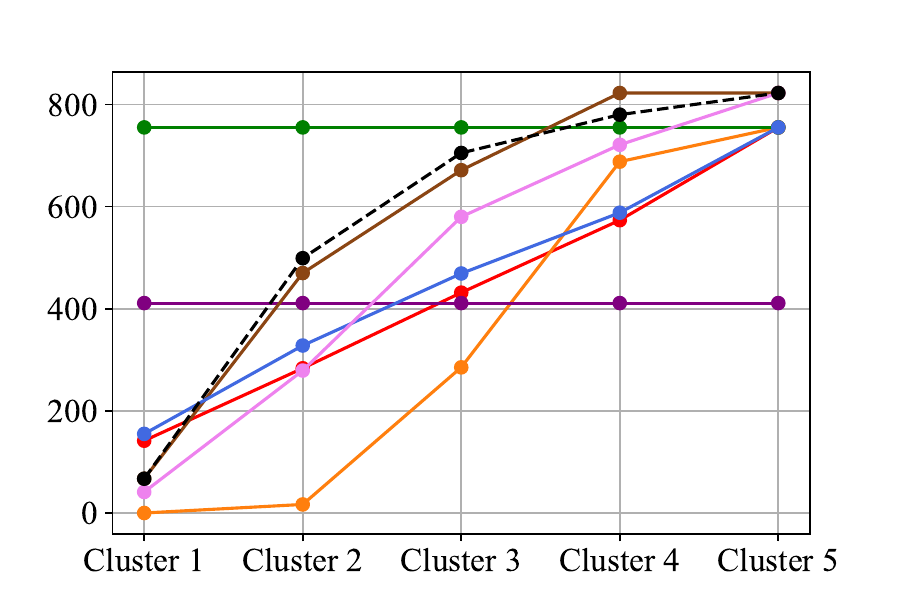}
    \end{subfigure}
    \begin{subfigure}[t]{0.24\textwidth}
        \centering
        ResNet44
        \includegraphics[width=\textwidth,
        ]{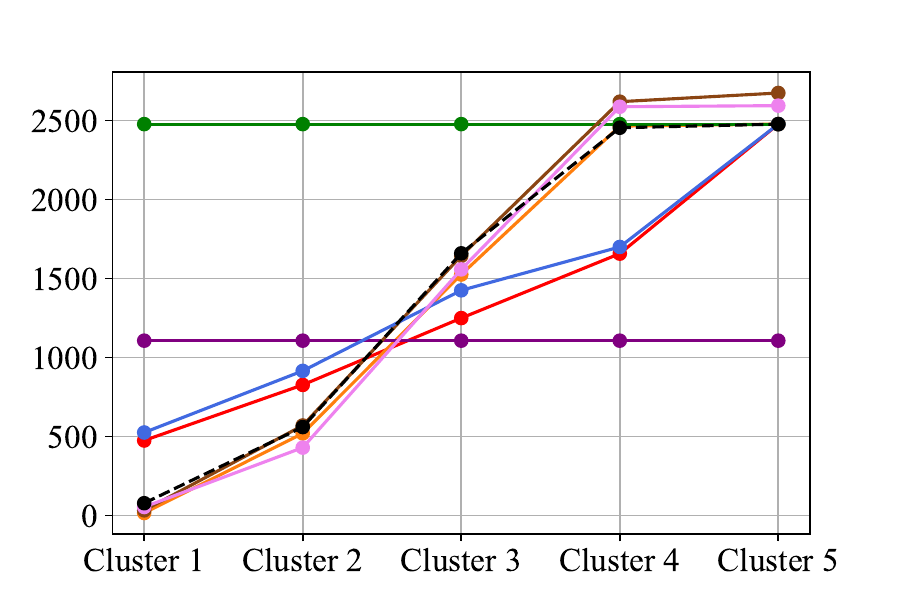}
    \end{subfigure}
    \caption{The theoretical memory consumption (MB) among clients.}
    \label{fig:theo_memory}
\end{figure}

\textbf{Practical and theoretical memory consumptions:} As shown in Figure \ref{fig:act_mem}, FedOLF effectively reduces the memory consumption practically (We merge the curves of Feddrop, FjORD, HeteroFL for brevity as their memory consumptions are very close). Moreover, considering that the real memory usage is usually context-dependent (physical device, programming language, etc), we also calculate their theoretical memory usage following \cite{slt}. As shown in Equation (\ref{eq:mem}), for a neural network \(w\) with \(N\) layers, the memory consumption \(m(w)\) can be computed as:
\begin{equation}\label{eq:mem}
\begin{split}
    m(w) & = \sum_{q=1}^{N} m_{\text{AM}}(W_{q})+m_{\text{G}}(W_{q})+m_{\text{W}}(W_{q}) \\
    & \approx \sum_{q=1}^{N} m_{\text{AM}} (W_{q})
\end{split}
\end{equation}
That is, the overall memory consumption \(m\) is the accumulated memory consumption of three components, which are parameter weights (\(m_{\text{W}}\)), gradients (\(m_{\text{G}}\)), and activation maps (\(m_{\text{AM}}\)) across all layers. Moreover, compared with weights and gradients, the size of activation maps is much larger and consumes a dominant memory space. Therefore, the overall memory consumption can be approximated as the total size of activation maps across all layers \cite{slt}. As shown in Figure \ref{fig:theo_memory}, FedOLF effectively alleviates the theoretical memory consumption as well. The reason is that, in FedOLF, for a frozen layer \(W_{q}\), \(m_{\text{AM}}(W_{q})\) becomes zero, as no activation maps have to be stored for training \cite{slt}. Accordingly, a client \(k\) can choose \(l_{k}\) (i.e., the number of frozen layers) to be the largest value, given \(\sum_{q=l_{k}+1}^{N} m_{\text{AM}}(W_{q})\) (the size of activation maps in the remaining active layers) not exceeding its memory limit. Moreover, in real scenarios where the memory space may vary, a client can dynamically adjust \(l_{k}\) based on the real-time memory limit. In the future, we aim to consolidate FedOLF with more comprehensive layer-freezing strategies considering more practical factors, such as computation resources, bandwidth, battery, and storage. \par

\begin{figure}[ht]
    \centering
     \begin{subfigure}[t]{0.24\textwidth}
        \centering
        \includegraphics[width=\textwidth,
        ]{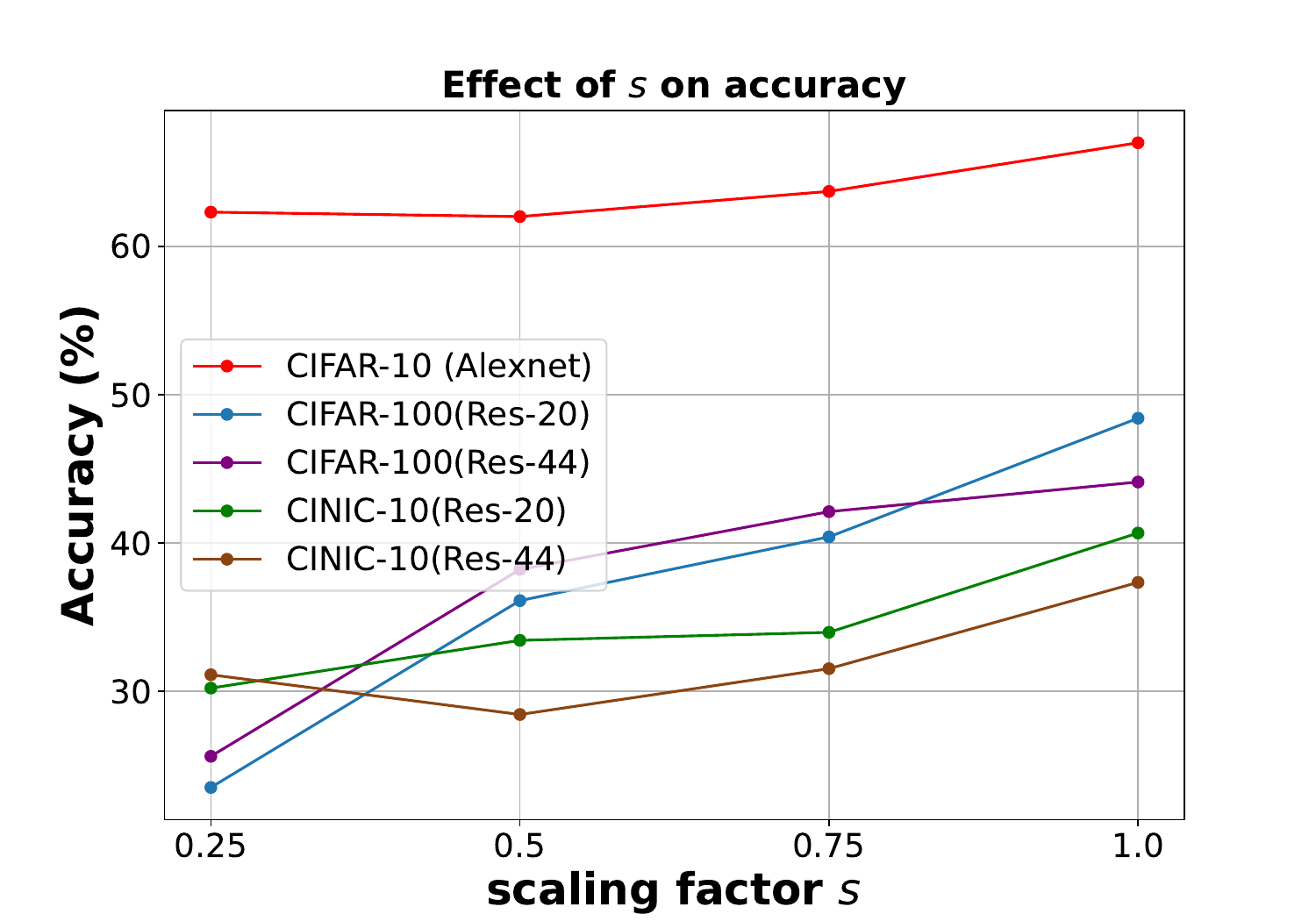}
        \caption{\(s\) vs. accuracy (iid).}
        \label{fig:stune_iid}
    \end{subfigure}
    \begin{subfigure}[t]{0.24\textwidth}
        \centering
        \includegraphics[width=\textwidth,
        ]{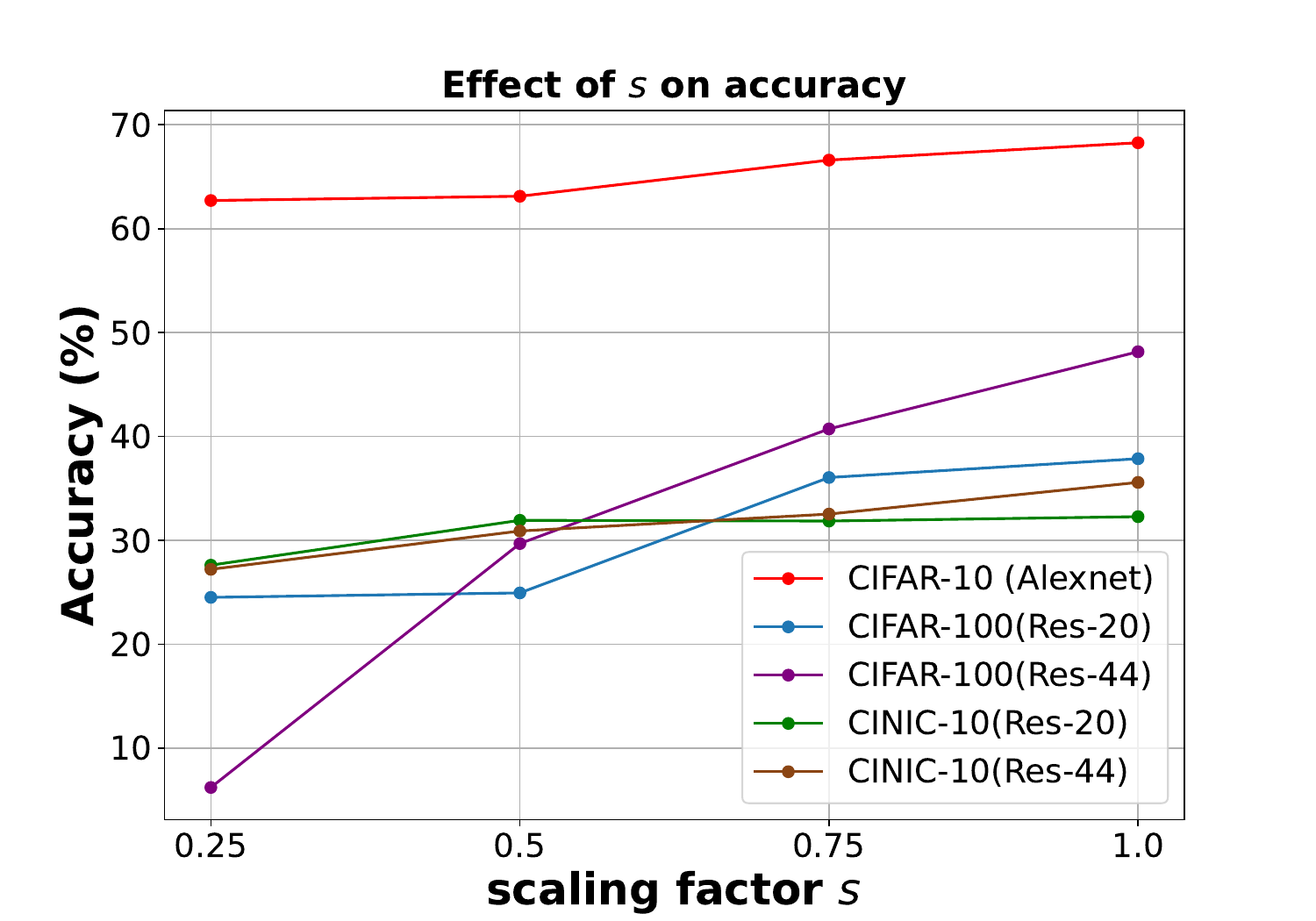}
        \caption{\(s\) vs. accuracy (non-iid).}
        \label{fig:stune_non-iid}
    \end{subfigure}
    \caption{Effect of the TOA scaling factor \(s\) on accuracy}\label{fig:stune}
    \centering
    \includegraphics[width=0.3\textwidth,
    ]{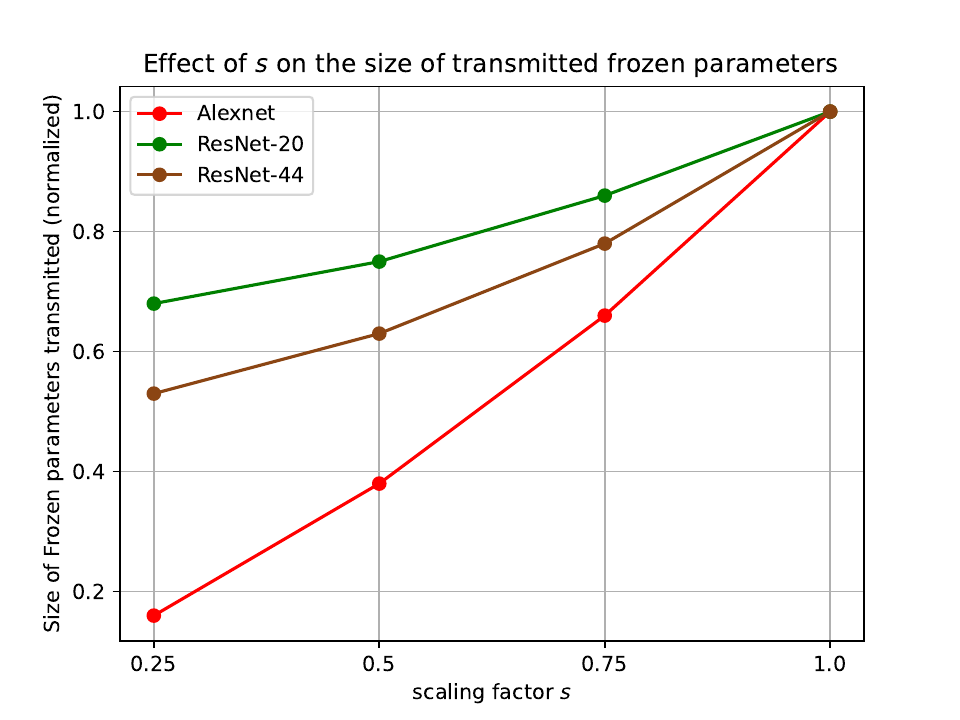}
    \caption{Effect of the TOA scaling factor \(s\) on the size of the frozen layers.}\label{fig:slf_tune}
\end{figure}

\begin{figure}[htbp]
    \centering
    \begin{subfigure}[t]{0.24\textwidth}
        \centering
        \includegraphics[width=\textwidth,
        ]{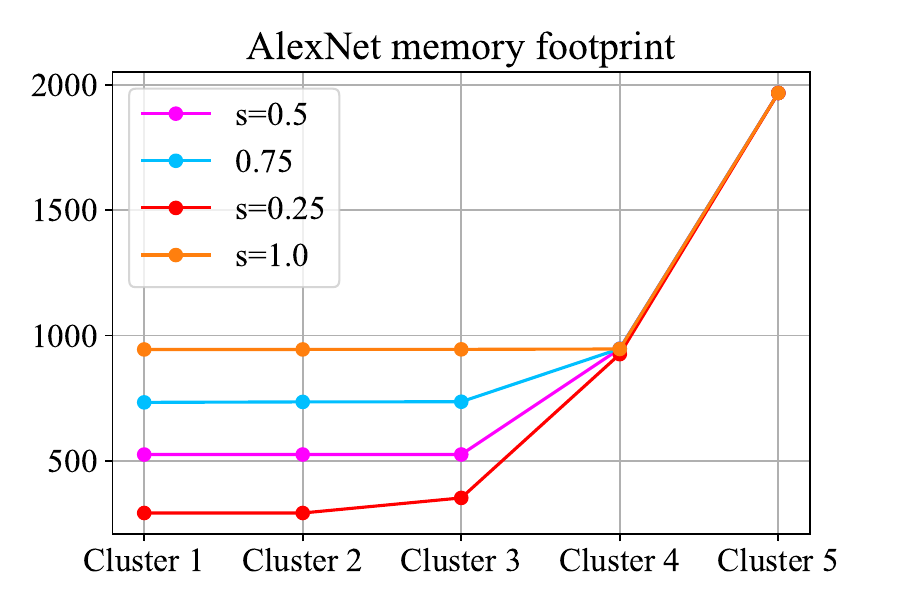}
        \caption{AlexNet.}
    \end{subfigure}
    \begin{subfigure}[t]{0.24\textwidth}
        \centering
        \includegraphics[width=\textwidth,
        ]{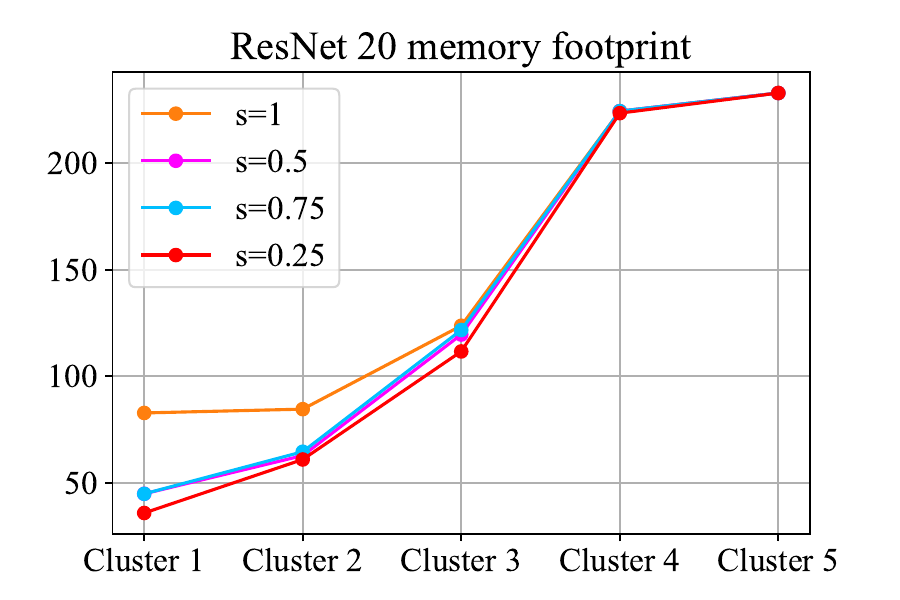}
        \caption{ResNet20.}
    \end{subfigure}
    \begin{subfigure}[t]{0.24\textwidth}
        \centering
        \includegraphics[width=\textwidth,
        ]{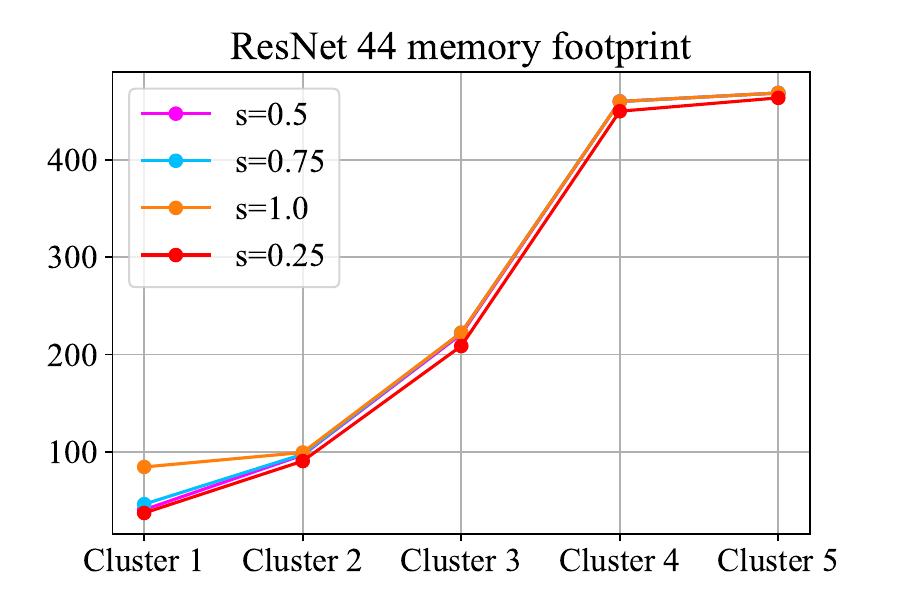}
        \caption{ResNet44.}
    \end{subfigure}
    \caption{Effect of the TOA scaling factor \(s\) on the practical memory consumption (MB).}
    \label{fig:toa_memory}
\end{figure}

\begin{figure}[htbp]
    \centering
    \includegraphics[width=0.4\textwidth,
        ]{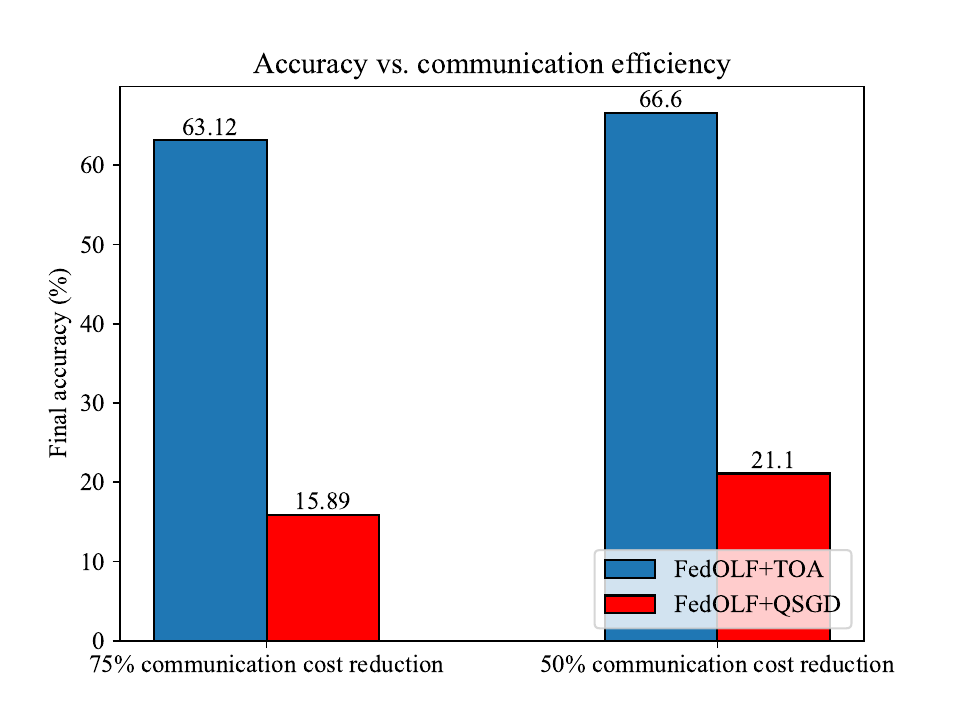}
    \caption{TOA vs. QSGD on the CIFAR-10 dataset.}
    \label{fig:toaqsgd}
\end{figure}

\textbf{Hyperparameter tuning and ablation study:} We tune the scaling factor of TOA \(s\) using a grid search within \(\{0.25, 0.5, 0.75, 1.0\}\), where \(s=1\) is equivalent to FedOLF without TOA. Results in Tables \ref{tab:acc_iid}, \ref{tab:acc} and Figures \ref{fig:stune} and \ref{fig:slf_tune} reveal that TOA effectively reduces the downstream communication cost without degrading much accuracy (except for CIFAR-100 with ResNet-44). For example, a scaling factor \(s=0.25\) can reduce the size of the transmitted frozen parameters by utmost 84\% with a minor 5.56\% accuracy loss compared with FedOLF sole (AlexNet). Besides, TOA further reduces the practical memory consumption as Figure \ref{fig:toa_memory} shows. Additionally, we compare TOA with the well-recognized quantized SGD (QSGD) method \cite{qsgd} for AlexNet on CIFAR-10. As shown in Figure \ref{fig:toaqsgd}, TOA achieves much higher accuracy than QSGD given the same degree of communication efficiency. Specifically, TOA (\(s=0.5\)) is compared with QSGD with 8 bits and  TOA (\(s=0.75\)) is compared with QSGD with 16 bits so that their reductions of communication cost are approximately equal. \par

\subsection{Results analysis}\label{subsec:discussion}
In summary, FedOLF significantly outperforms existing resource-constrained FL works by obtaining the highest accuracy on all datasets and models. Besides, the accuracy of FedOLF is only marginally below the standard FedAvg framework (Tables \ref{tab:acc_iid} and \ref{tab:acc}), indicating the ability of FedOLF in training a near-optimal model in resource-constrained settings. We attribute this result to two key components in the design of FedOLF, which are bounded gradient error and low-level later sharing, as discussed in Section \ref{subsec:fedolf}. Subsequently, FedOLF significantly increases the energy efficiency by achieving the highest accuracy with the same energy consumption as \ref{fig:overall_efficiency} shows. Moreover, as shown in Figures \ref{fig:act_mem} and \ref{alg:toa}, FedOLF effectively reduces the memory consumption due to the efficacy of the OLF module in alleviating memory usage (see Figure \ref{lf_memory}). Furthermore, we consolidate FedOLF with TOA to further reduces the energy and memory consumption while maintaining accuracy (see Figures \ref{fig:overall_cost} and \ref{fig:toa_memory}). \par

\begin{figure}
    \centering
    \includegraphics[width=\linewidth]{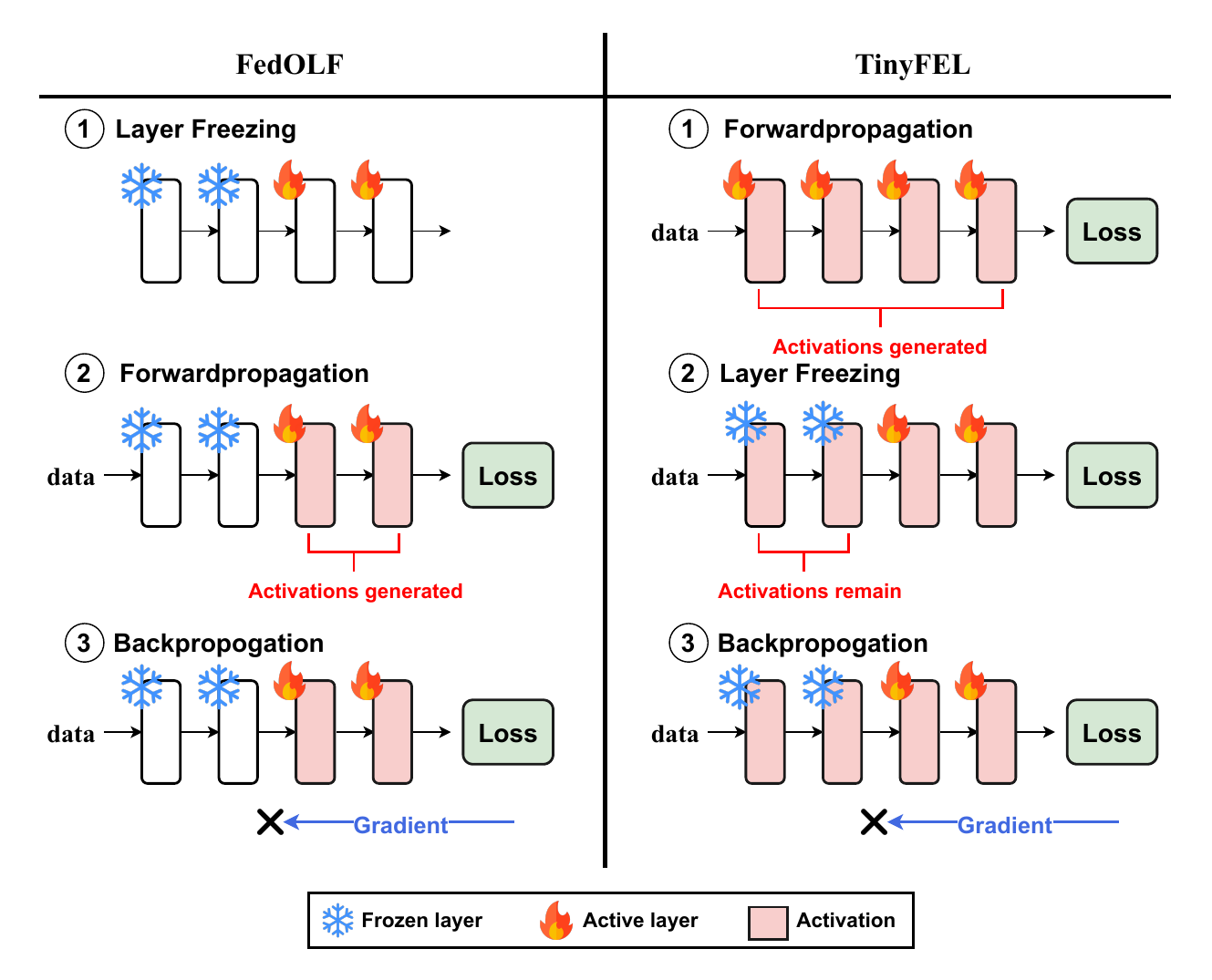}
    \caption{The difference between FedOLF and TinyFEL depends on the timing of layer freezing.}
    \label{fig:tinyfelcompare}
\end{figure}

\subsection{Distinction between FedOLF and TinyFEL}\label{sec:tinyfel}
Although the design of FedOLF resembles the previous TinyFEL \cite{tinyfel} work by freezing low-level layers in training, FedOLF differs from TinyFEL in the timing of layer freezing. As depicted in Figure \ref{fig:tinyfelcompare}, in FedOLF, low-level layers are frozen before local training starts. This module ensures that the frozen layers will not generate any activations during training, which significantly alleviates the memory consumption. \par
Comparatively, in TinyFEL, low-level layers are frozen in backpropagation while remaining active in forward propagation. However, for practical implementations like Pytorch, activations are generated in forward propagation as Figure \ref{fig:tinyfelcompare} shows. The reason is that in Pytorch, all active layers operate as an entire computational graph rather than individual modules, and produce activations simultaneously in the forward pass. In this scenario, although the low-level layers are frozen during backpropagation, they still account for a large memory space owing to the massive magnitude of remaining activations. \par

\begin{figure}[ht]
    \centering
    \includegraphics[width=0.8\linewidth]{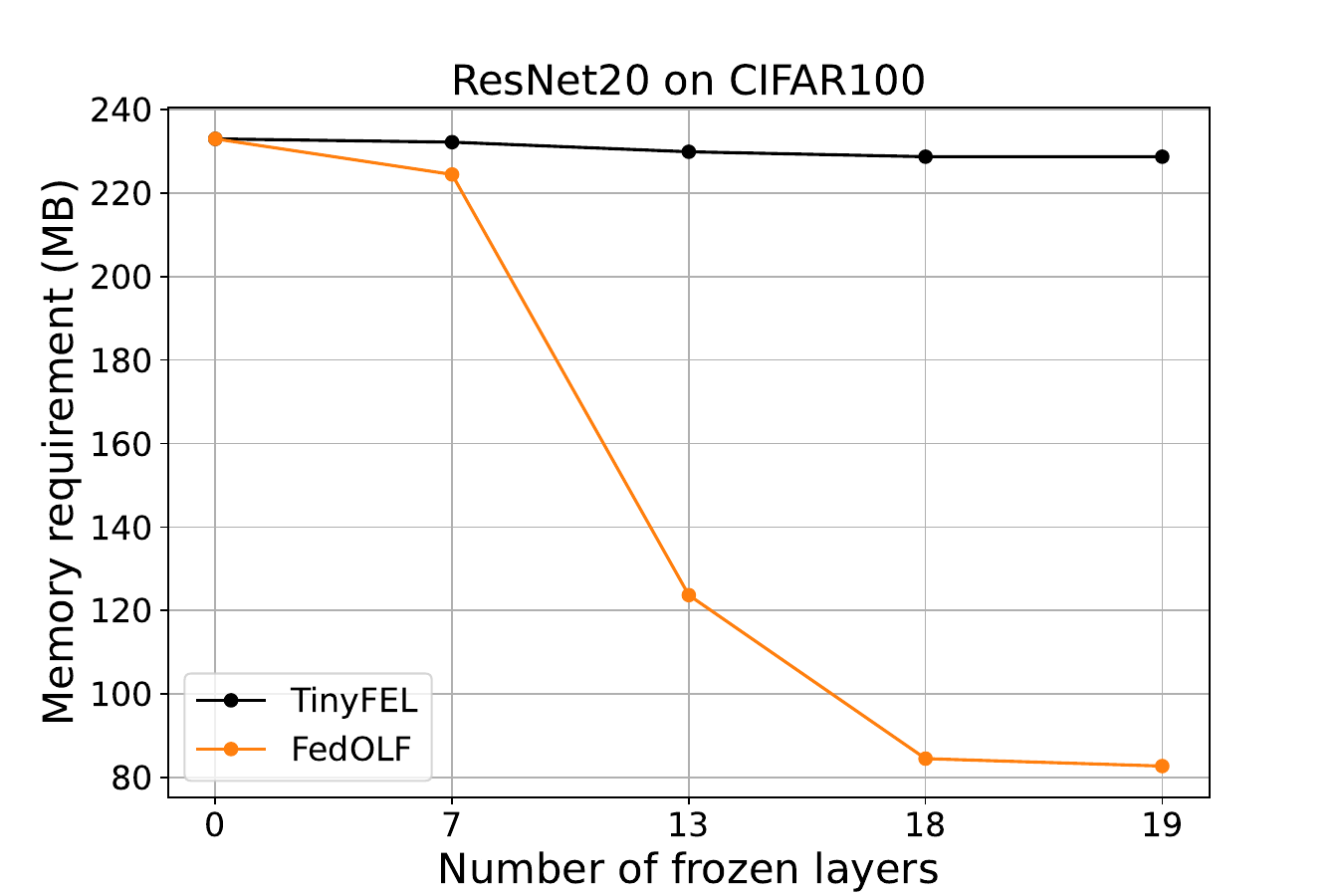}
    \caption{The realistic memory consumption of FedOLF and TinyFEL on CIFAR-100 with ResNet20.}
    \label{fig:tinyfelmem}
\end{figure}

For verification, we evaluate the actual memory consumption of FedOLF and TinyFEL on CIFAR-100 with ResNet20 using {\fontfamily{qcr}\selectfont TORCH.CUDA.MAX\underline{\hspace{1mm}}MEMORY\underline{\hspace{1mm}}ALLOCATED} function. As Figure \ref{fig:tinyfelmem} shows, TinyFEL consumes significantly more memory than FedOLF, demonstrating its practical memory inefficiency.
